\documentclass{article}

\usepackage{microtype}
\usepackage{graphicx}
\usepackage{subfigure}
\usepackage{booktabs} 
\usepackage{epigraph}
\usepackage{enumitem}
\setlist{nosep}

\usepackage{hyperref}

\usepackage[accepted]{icml2024}

\usepackage{amsmath}
\usepackage{amssymb}
\usepackage{mathtools}
\usepackage{amsthm}

\usepackage[capitalize,noabbrev]{cleveref}

\theoremstyle{plain}

\theoremstyle{definition}

\theoremstyle{remark}

\usepackage[textsize=tiny]{todonotes}

\icmltitlerunning{LLMs as Method Actors}

\begin{document}

\twocolumn[
\icmltitle{LLMs as Method Actors:\\
A Model for Prompt Engineering and Architecture}



\icmlsetsymbol{equal}{*}

\begin{icmlauthorlist}
\icmlauthor{Colin Doyle}{sch}
\end{icmlauthorlist}

\icmlaffiliation{sch}{Loyola Law School, Loyola Marymount University, Los Angeles, CA, USA}

\icmlcorrespondingauthor{Colin Doyle}{colin.doyle@lls.edu}

\icmlkeywords{LLMs, Large Language Models, Connections Puzzles, Puzzles, Word Puzzles, Word Games, Reasoning, Complex Reasoning, Prompt Engineering, Prompt Architecture, Actors, Acting, Method Actors, GPT-4o, o1, o1-preview}

\vskip 0.3in
]



\printAffiliationsAndNotice{}  

\begin{abstract}
We introduce ``Method Actors'' as a mental model for guiding LLM prompt engineering and prompt architecture. Under this mental model, LLMs should be thought of as actors; prompts as scripts and cues; and LLM responses as performances. We apply this mental model to the task of improving LLM performance at playing Connections, a New York Times word puzzle game that prior research identified as a challenging benchmark for evaluating LLM reasoning. Our experiments with GPT-4o show that a ``Method Actors'' approach can significantly improve LLM performance over both a vanilla and ``Chain of Thoughts'' approach. A vanilla approach solves 27\% of Connections puzzles in our dataset and a ``Chain of Thoughts'' approach solves 41\% of puzzles, whereas our strongest ``Method Actor'' approach solves 86\% of puzzles. We also test OpenAI’s newest model designed specifically for complex reasoning tasks, o1-preview. When asked to solve a puzzle all at once, o1-preview solves 79\% of Connections puzzles in our dataset, and when allowed to build puzzle solutions one guess at a time over multiple API calls, o1-preview solves 100\% of the puzzles. Incorporating a ``Method Actor'' prompt architecture increases the percentage of puzzles that o1-preview solves perfectly from 76\% to 87\%.
\end{abstract}

\section{Introduction}
\label{Introduction}

We introduce ``Method Actors'' as a mental model for guiding LLM prompt engineering and prompt architecture. Under this mental model, LLMs should be thought of as actors; prompts as scripts and cues; and LLM responses as performances. Four principles for prompt writing and task decomposition follow from this mental model: 1) Prompt engineering is playwriting and directing. 2) Performance requires preparation. 3) Complex tasks should be decomposed to the point at which imitation and authenticity produce equivalent results. 4) Where imitation fails, compensate with methods that do not rely upon LLMs.

We apply this mental model to the task of improving LLM performance at playing Connections, a New York Times word puzzle game that prior research has identified as a useful benchmark for testing LLM complex reasoning performance. \cite{samadarshi2024,todd2024}. With these puzzles, a player is shown a four-by-four grid of 16 words and must identify four groups of four words that have a unique connection to one another. Each word appears in exactly one of the four groups, and each of the four groups is unique. For each game, the player is allowed to make up to three incorrect guesses and still solve the puzzle. Connections puzzles test a variety of reasoning skills for a player,  and the open-ended and qualitative nature of Connections puzzles makes solving these puzzles a unique challenge for LLMs. The New York Times reports that, ``In Connections, there’s no way to use math or even artificial intelligence to reliably solve the game.''\footnote{https://www.nytimes.com/2024/08/27/upshot/connections-bot-faq.html}

Our experiments with GPT-4o show that a ``Method Actors'' approach can significantly improve LLM performance over both a vanilla and ``Chain of Thoughts'' approach. A vanilla approach solves 27\% of the Connections puzzles in our dataset and solves 12\% perfectly without making an incorrect guess. A ``Chain of Thoughts'' approach solves 41\% of the puzzles and solves 20\% perfectly. Our initial ``Method Actor'' approach solves 78\% of the puzzles and solves 41\% perfectly. Our revised ``Method Actor'' approach solves 86\% of the puzzles and solves 50\% perfectly.

After these initial experiments were performed, OpenAI released o1-preview, an LLM model that specializes in performing complex reasoning tasks like Connections puzzles.\footnote{https://openai.com/o1/} Further experiments with o1-preview reveal o1-preview’s superior baseline performance at Connections puzzles over GPT-4o and also reveal that using the ``Method Actor'' mental model can improve the rate at which o1-preview solves puzzles perfectly. A ``one shot'' approach — in which o1-preview must solve the puzzle with one prompt and response — solves 79\% of the puzzles and solves 72\% perfectly. When allowed to build puzzle solutions one guess at a time over multiple API calls, a vanilla approach solves 100\% of the Connections puzzles and solves 76\% perfectly. A ``Method Actor'' approach solves 99\% of the puzzles and solves 87\% perfectly.

Based upon prior evaluations of human performance at solving connections puzzles perfectly, the ``Method Actor'' approaches with GPT-4o perform better than human novices and slightly worse than human experts. \cite{samadarshi2024} Each of the o1-preview approaches surpasses human expert performance at solving puzzles perfectly, with the ``Method Actor'' approach having the strongest performance. A comparison of LLM performance measured against the New York Times’ assessment of puzzle difficulty reveals that puzzles that are easy for people also tend to be easy for LLMs, and puzzles that are difficult for people also tend to be difficult for LLMs.

This paper demonstrates how using ``Method Actors'' as a mental model for LLMs can improve LLM performance at Connections puzzle, but leaves open the opportunity to examine how this mental model affects LLM performance with other tasks. Our code and data is available to the public at https://github.com/colindoyle0000/llms-as-method-actors.

\section{Related Work}
\label{Related Work}

\subsection{Prompt Engineering and LLM System Architecture}
In recent years, large language models have shown a remarkable improvement in performance on natural language processing tasks. \cite{chang2023b}. A wave of contemporary research focuses on improving large language models’ performance at complex tasks through novel prompting techniques. These techniques can encompass both prompt engineering and prompt architecture. Prompt engineering refers to techniques for writing individual prompts to elicit LLM responses that exhibit stronger performance at natural language tasks. Prompt architecture refers to methods for structuring multiple prompts and responses to better accomplish complex tasks through a series of LLM calls. In contrast to conventional methods for improving LLM performance, prompting techniques do not require extensive retraining or fine-tuning, making these methods both cost-effective and widely accessible. \cite{vatsal2024}. Prompting techniques such as Chain-of-Thought, Tree-of-Thoughts, ReAct, and Self-consistency with CoT have shown to improve LLM performance at different complex reasoning tasks. \cite{wei2023, yao2023b, yao2023a, wang2023d}. Chain-of-Thought (CoT) is a prompt engineering technique for guiding an LLM to develop its reasoning step-by-step before reaching an ultimate conclusion. \cite{wei2023} Tree of Thoughts (ToT) builds upon CoT with the prompt architecture of a thought tree with multiple reasoning branches. \cite{yao2023b}. But even state-of-the-art methods are limited. Their performance gains are limited to specific examples, and ``concepts related to the LLM reasoning are not well-defined, hindering effective design of new more powerful schemes.'' \cite{besta2024}. Unexpected — and seemingly inexplicable — results continue to crop up, such as research that found that injecting emotional stimuli into prompts such as, ``This is very important to my career,'' can improve LLM reasoning performance. \cite{li2023e}. The goal of many of these techniques is to engineer a process by which an LLM mimics the steps of multi-step thinking and reasoning that humans perform. But ``while many schemes rely on the notion of the LLM thought, it is not clear how it relates to concepts such as a prompt.'' \cite{besta2024}. Some research conceptualizes LLM responses not as ``thoughts'' but as imitations of the products of thought. \cite{banerjee2024, ferrucci2010}.

\subsection{Connections Puzzles as a Reasoning Benchmark}

The advent of large language models has led many researchers to evaluate the models' performance playing and generating text-based reasoning games. \cite{gallotta2024}. Prior work has evaluated LLM performance at NPR word puzzles; crossword puzzles; and the popular language-based board game, Codenames, among others. \cite{jaramillo2020,zhao2023a,yao2023b}. Another line of research has evaluated LLM performance at generating text-based puzzles, including Connections puzzles. \cite{merino2024}. Throughout the history of A.I., games have been used as benchmarks for model performance, including famous examples of chess, go, and Jeopardy. \cite{bory2019}.

Connections Puzzles have been proposed as a useful benchmark for testing LLM complex reasoning performance. \cite{samadarshi2024, todd2024}. Connections puzzles test a variety of reasoning skills for a player, including the ability to flexibly switch between methods of reasoning to solve a puzzle. The open-ended and qualitative nature of Connections puzzles makes solving these puzzles a unique challenge. Solving connections puzzles is a seemingly impossible task for LLMs, as even the New York Times reports that, ``In Connections, there’s no way to use math or even artificial intelligence to reliably solve the game.''\footnote{https://www.nytimes.com/2024/08/27/upshot/connections-bot-faq.html} At the same time, because Connections puzzles have clearly specified correct and incorrect answers, they provide objective criteria for evaluating an LLM system's open-ended reasoning abilities.

Prior work has found that LLMs can solve some Connections puzzles but that there is much room for improvement. \citet{todd2024} found that GPT-4o could solve 38.93\% of Connections puzzles within an python-based system that used chain-of-thought prompting, used separate API calls to have the model to submit guesses one at a time, and gave the model feedback about whether its guesses were correct or incorrect. \citet{samadarshi2024} found that GPT-4o could solve only 8\% of puzzles when asked to solve the puzzle completely with one chain-of-thought prompt and response. In evaluating LLM performance, \citet{samadarshi2024} created a taxonomy of reasoning skills and knowledge required to solve Connections puzzles, noting where LLMs succeed and fall short. 

Performance at Connections puzzles may be guiding the development of LLM models as well. While developing a specialized complex reasoning model, o1-preview, OpenAI employees reportedly used the model's performance on Connections puzzles as proof of the model's performance at complex reasoning tasks. \cite{lambert2024}.

This paper builds on that research by evaluating the performance of GPT-4o using “Method Actors” as a mental model for guiding LLM prompt engineering and prompt architecture. The paper also evaluates the performance of OpenAI's new o1-preview model, both with standard prompting and with a “Method Actor” prompt architecture.

\section{Method Acting with Large Language Models}
\label{mmop}

\epigraph{
All the web’s a stage,

And all the models merely mimics;  

They have their prompts and their tokens,  

And one model in its code plays many roles,  

Its scripts being endless acts.}
{Chat-GPT-4o 

(response to a prompt asking it to rewrite Shakespeare lines to be about LLMs as method actors)}

We introduce ``Method Actors'' as a mental model for LLMs that can guide prompt engineering and prompt architecture. Under this mental model, LLMs should be thought of as actors; prompts as scripts and cues; and LLM responses as performances. LLMs and actors have a lot in common. Both mimic the product of human thought and emotion. Success for both actors' and LLMs' performance is often measured by a performance's verisimilitude: how much the imitation of human thought and feeling seems authentic. \cite{chiang2024}. Hallucinations are the sin qua non of both actors and LLMs. Actors' performances are faithful to the text of a script but not to external reality, just as LLMs' responses are faithful to the text of a prompt but not the external truth of the world. The analogy of LLMs as actors is more than cute — it's useful. Imagining LLMs as actors performing a part can better align a user's expectations with LLMs' capabilities because prompting is more like giving a performer a cue than asking a robot mind for its thoughts. Four principles for prompt writing and task decomposition follow from this mental model:

\paragraph{Prompt engineering is playwriting and directing.} Prompts should set the scene like a playwright would: providing the LLM with a character, motivation, setting, and stage direction. Beyond just assigning the LLM a role, the scene should provide motivation and direction, setting up a story for the LLM to perform. Direct instructions for an LLM are like stage directions for an actor. They should focus on the form of the LLM’s response, delineating the steps for an LLM to follow. Improvisation can be channeled effectively by scripting out the beats of the LLM’s performance and the patterns of language to use.

\paragraph{Performance requires preparation.} Under this mental model, LLM responses are not understood as \textit{thoughts} but as \textit{performances}. Just as the verisimilitude of an acting performance depends upon off-screen preparation — the many smaller, unseen steps an actor takes to build up to the present mental state of the character — the verisimilitude of an LLM's performance often requires similar background preparation. LLMs imitate the products of thinking, not thinking itself. For a complex performance, an LLM should be prompted to produce the products of any ``behind the scenes'' thinking required for a complex task. These intermediary performances can build incrementally to culminate in a complex final performance. To manage an LLM’s context window, this often requires separate API calls that feed information forward to subsequent LLM calls.

\paragraph{Complex tasks should be decomposed to the point at which imitation and authenticity produce equivalent results.} When accurate results are required, a prompt architecture should be designed so that it relies not upon the truthfulness of an LLM's response but rather upon the authenticity of the LLM's performance. LLM responses are more reliable and accurate when an imitation of performing an action is equivalent to genuinely performing an action. The inverse is true. LLMs are less reliable and more prone to error when an imitation of a performing an action and genuinely performing an action produce divergent results. Therefore, a complex task should be decomposed into subtasks in which the imitation of performing the subtask and genuinely performing the subtask are equivalent. 

\paragraph{Where imitation fails, compensate with methods that do not rely upon LLMs.} Some complex tasks cannot be broken down into subtasks in which the imitation of performing the task and genuinely performing the tasks are equivalent. When this occurs, the broader LLM system or pipeline should be designed to compensate for potential errors through other programming methods.

\section{Experimental Setup}
\subsection{Game Details}

Connections Puzzles are an interactive word puzzle game published on the New York Times website and mobile games app.\footnote{https://www.nytimes.com/games/connections} The game debuted in June, 2023 and has quickly become one of the most popular and challenging puzzle games that the Times publishes. \cite{jennings2024, schwedel2024}. With these puzzles, a player is shown a four-by-four grid of 16 words and must identify four groups of four words that have a unique connection to one another. Each word appears in exactly one of the four groups, and each of the four groups is unique.

\begin{table}
    \centering
    \begin{tabular}{|c|c|c|c|} \hline 
         BOO-BOO&  GAFFE&  STING& POLLINATE\\ \hline
         DANCE&  LOU&  MICKEY& BUZZ\\ \hline
         PETUNIA&  JASMINE&  YOGI& FLUB\\ \hline
         POPPY&  BABE&  DAISY& GOOF\\ \hline
    \end{tabular}
    \caption{Connections Puzzle \#430}
    \label{tab:puzzle430}
\end{table}

The rules for Connections are simple, but the puzzles are challenging. Players must use a variety of reasoning methods and draw upon different sources of knowledge. \cite{samadarshi2024}. For each word, players must consider different possible meanings, relationships to words outside the puzzle, and unique linguistic attributes. The challenge of a Connections puzzle goes beyond brainstorming attributes for each word and lies in successfully grouping words together under a set of shared attributes. Table \ref{tab:puzzle430} shows the Connections puzzle \#430 for August 14, 2024. If you’re not already familiar with Connections puzzles, give this puzzle a try before you read ahead. By attempting to solve this puzzle, you’ll get a better sense of Connections puzzles’ level of difficulty and the reasoning skills required. See if you can identify four groups of four words that have a connection to one another. For the full interactive experience, you can do the puzzle online here: \href{https://connections.swellgarfo.com/nyt/430}{https://connections.swellgarfo.com/nyt/430}.

Connections puzzles are designed with a difficulty scale in mind. The Times labels each of the correct connections with a color — yellow, green, blue, or purple — representing the difficulty of identifying that connection. Each puzzle has one connection corresponding to each color. The connections for the easiest categories, yellow and green, are often words that are synonyms or words that belong to the same general category. For this puzzle, the yellow connection is ``BLUNDER'' and is comprised of the synonyms ``BOO-BOO,'' ``FLUB,'' ``GAFFE,'' and ``GOOF.'' The green connection is ``FLOWERS'' and includes different types of flowers: ``DAISY,'' ``JASMINE,'' ``PETUNIA,'' and ``POPPY.'' The more challenging connections typically require more precise knowledge or more creative thinking. The blue connection here is ``THINGS BEES DO,'' which are ``BUZZ,'' ``DANCE,'' ``POLLINATE,'' and ``STING.'' The purple connection is ``FIRST NAMES OF YANKEES LEGENDS'': ``LOU,'' ``BABE,'' ``YOGI,'' and ``MICKEY.'' Connections puzzles often include red herrings: connections between words that are not a part of the solution to the puzzle. In this puzzle, one red herring would be ``DISNEY CHARACTERS,'' which could include the words ``MICKEY,'' ``DAISY,'' and ``BUZZ.'' Another red herring could be ``CARTOON CHARACTERS,'' which could include ``BOO-BOO,'' ``YOGI,'' ``MICKEY,'' and ``DAISY.''

When playing Connections, users submit guesses one-by-one by selecting four words and clicking submit. For each game, the player is allowed to make up to three incorrect guesses and still solve the puzzle. Once the player has made either four correct guesses or four incorrect guesses, the game ends.

\subsection{Data}

Connections puzzles are released on a daily basis. This experiment used 100 recent New York Times Connections puzzles, 331 through 430, which were released between May 7, 2024 and August 13, 2024. The data for each puzzle, including the words and the solution, was acquired from a third-party website that collects past and present New York Times Connections puzzle information.\footnote{\href{https://connections.swellgarfo.com/archive}{https://connections.swellgarfo.com/archive}}

\section{Experiment: GPT-4o}

The first experiment compares the performance of different approaches for solving Connections puzzles using OpenAI's GPT-4o model: Vanilla, Chain-of-Thought, Chain-of-Thought (Scripted), Actor, and Actor-2. The first three approaches vary only in the natural language content of the prompts. The two Actor approaches vary both in the natural language content of the prompts and in the prompt architecture.

Some features are common to all five approaches. The LLM is asked to make and submit guesses one at a time. If the guess is a valid guess,\footnote{Valid guesses contains four words that are part of the list of words remaining to solve in the puzzle and are not identical to a set of four words of an incorrect guess that was already submitted. For each approach, the LLM system is prevented from submitted an invalid guess. If the guess to be submitted is not formatted correctly to be a set of four words remaining in the puzzle, the guess selection process starts over. If the guess to be submitted is a guess that was previously submitted and determined to be incorrect, the incorrect guess is not submitted a second time, and the guess selection process starts over.} the guess is submitted and checked against the puzzle solution. If the guess is correct, those four words are removed from consideration for subsequent guesses. If the guess is incorrect, that guess is saved as an incorrect guess that the LLM system cannot submit again. If the system has already submitted any incorrect guesses, a list of guesses known to be incorrect is included in subsequent prompts. Upon making three correct guesses, the system automatically submits the fourth guess because — by process of elimination — it is the only possible guess that remains, even if the LLM has not identified the connection that those four words share. The process ends when the system has submitted either four correct or four incorrect guesses.

\subsection{Vanilla}

The Vanilla approach sets a baseline for how GPT-4o performs at solving connections puzzles. Each guess consists of two separate API calls to the LLM. The first call prompts the LLM to guess one four-word group that represents part of the solution to the puzzle. The prompt is an edited version of the instructions on The New York Times Connections website that provides the LLM with more details about the puzzle than the website instructions provide. If the LLM cannot come up with a good guess, it is allowed to not submit a guess.\footnote{The prompt includes the instruction, ``If you still can't identify a guess to submit, say 'I can't identify a good guess to submit.''' When this occurred, the system would not submit a guess and would restart the process of generating a guess with a new API call.} The second call takes the LLM’s response to the first prompt and asks the LLM to extract the guess information and format the guess in a way for subsequent steps to parse and submit. 

\subsection{Chain-of-Thought}

The Chain-of-Thought approach is identical to the Vanilla approach except for changes to the ``make a guess'' prompt. In this prompt, the LLM is assigned a role, ``You are a professional puzzle solver.'' and at the end of the prompt the LLM is instructed,``Let’s think this through step-by-step.'' This now-ubiquitous closing instruction has been demonstrated many times over to improve LLM reasoning performance by having the LLM work through reasoning steps 'aloud' in the output of its responses. \cite{wei2023, kojima2023}.

\subsection{Chain-of-Thought (Scripted)}

The Chain-of-Thought (Scripted) approach is identical to the Chain-of-Thought approach except for two changes to the ``make a guess'' prompt. First, the prompt includes a more expansive, carefully curated set of examples of puzzle solutions. Second, the prompt instructs the LLM to approach the puzzle solving task following a particular set of steps.

The prompt includes twenty-three examples of correct guesses to Connections puzzles. By comparison, the Vanilla and Chain-of-Thought approaches only had two examples of correct guesses (taken from the instructions on the New York Times website). The examples were chosen to represent each of the types of answers seen in Connections puzzles.\footnote{These templates were drawn from real Connections puzzles from December 2023 and March 2024.}

The prompt includes instructions on a two-step process for the LLM to follow. The first step is to identify two words that have a connection with each other. The second step is to look through the remaining words and see if other words share the same connection. The LLM is also instructed to start over at the first step if it is unable to identify a group of four words that share a connection. These instructions were chosen to structure the LLM’s thinking to follow a method commonly used by people solving Connections puzzles.

\subsection{Actor}

The Actor approach was designed according to the principles articulated in Section \ref{mmop}: prompt engineering is playwriting and directing; performance requires preparation; complex tasks should be decomposed to the point at which imitation and authenticity produce equivalent results; and where imitation fails, compensate with methods that do not rely upon LLMs.

The prompts for this approach include dramatic scene setting and role definition. The prompts inform the LLM that it is a professional puzzle solver who has been brought in by the FBI because terrorists have planted a bomb inside a children’s hospital, and the only way to defuse the bomb is by solving this word puzzle correctly. An informal analysis comparing the LLM responses from prompts that either used or didn't use this scene-setting text revealed that including this scene-setting text resulted in the LLM using more of its output context window to keep thinking through possible solutions and less frequently concluding that it could not find an answer.

For prompt architecture, the task of solving Connections puzzles is divided into two stages: brainstorming and discernment. The brainstorming stage consists of five separate prompts for the LLM to brainstorm potential guesses. Because the answers to Connections puzzles follow particular patterns, the LLM’s task is framed to imitate these past performances. The process cycles through a set of 24 templates representing the patterns of answers featured in Connections puzzles.\footnote{Like the examples used in the Chain-of-Thought (Scripted) approach, these templates were drawn from real Connections puzzles from December 2023 and March 2024.} For each brainstorming call, one of the 24 templates is selected and the LLM is prompted to generate a guess by applying that template to the word list for the current puzzle. Each template includes a description of a particular pattern of an answer along with examples of that pattern from prior puzzles. For example, the LLM may be asked to identify ``words that can be synonymous adjectives with each other'' or ``words that share a pop culture reference'' or ``words that are each followed by the same word or phrase.'' The templates include a script of multi-step instructions on the process for the LLM to follow. 

After brainstorming guesses, discernment begins. Discernment is more of a challenge for decomposition than brainstorming. With brainstorming guesses, authentically brainstorming potential guesses is more-or-less equivalent to the imitation of brainstorming potential guesses, particularly when the imitation is accomplished through applying the patterns of past answers to new words. The task of discernment involves judgment and decision-making about a unique set of possible guesses. Brainstorming can be scripted for an actor in a way that discernment cannot. Because an imitation of discernment is unlikely to be equivalent to discernment, the prompt architecture needs to compensate for potential weaknesses. This is accomplished by giving the LLM multiple passes to winnow down potential guesses and by providing discernment scripts with criteria for the LLM to apply — as much as it is possible to do so. 

The discernment process begins with an ``extract'' stage to reduce the quantity of information for the LLM to process to only include viable guesses. The LLM is called to extract from the brainstorming notes any valid potential guesses and the rationale behind those guesses. This information is passed on to the ``discern'' stage in which the LLM is asked to discern the strongest guess from among the possible guesses. Lastly, at a ``decide'' stage, the LLM is prompted to decide whether the process has produced a guess worth submitting. Guesses worth submitting are stockpiled. Once five guesses are ready to submit, a final ``evaluation'' stage begins.\footnote{After two correct guesses have been submitted, the evaluation process starts once three guesses are ready to submit.} The LLM is asked to consider the potential guesses and select the strongest one to submit. This process is similar to the ``Tree of Thoughts'' framework and other approaches that involve generating ideas and then iteratively pruning down possibilities until only the strongest options remain. \cite{yao2023b, yao2024}.

\subsection{Actor-2}

A second ``method acting'' approach, Actor-2, was conducted to test additional methods for working around LLM shortcomings within the Method-Actor approach. A qualitative analysis of the LLM responses at each stage in the Actor process revealed that the weak point in the process was the discernment stage. The LLM was able to successfully brainstorm many different possible guesses but struggled to distinguish between the good and bad guesses.

Under the principles spelled out in Section \ref{mmop}, options for revision include further decomposition or engineering around the LLM's shortcomings. Given that further decomposition did not seem viable, Actor-2 instead takes the LLM's weakness at discernment as a given and attempts to engineer the broader system to compensate for this weakness. Two changes are introduced. First, external validation criteria must be satisfied before a guess that the LLM has chosen to submit is actually submitted. Second, a validation process is introduced to filter out hallucinations.

Actor-2 keeps the brainstorm and discern processes the same as the Actor approach but revises the evaluate process to no longer rely exclusively upon the LLM's discernment over which potential guesses are strongest. This approach uses deterministic logic to help the system navigate around red herrings, which LLMs struggle to identify. After the LLM chooses to submit a guess, the guess is added to a ``final guesses'' list instead of submitting it right away. The system analyzes — without an LLM’s involvement — whether any guesses in the ``final guesses'' list form unique pairs (i.e., there is no overlap of words between the two guesses). Once the ``final guesses'' list includes two guesses that are unique pairs, the guesses are submitted. In this way, the system deterministically reduces the frequency of submitting red herrings. Alternately, if the same guess appears in the ``final guesses'' list three times, the guess is submitted. One potential problem with the ``final guesses'' list is that it risks making the system much less efficient. The system may produce the same guess over and over again. To ensure that the LLM produces a diversity of guesses and explores the larger puzzle space of possibilities, words are sometimes removed from the possible words list when generating guesses. The removed words are words from guesses in the ``final guesses'' list and words from guesses that are waiting to be evaluated.

Once the system has submitted at least two incorrect guesses, the discern stage includes a validation process for filtering out hallucinations. If a proportion of hallucinated guesses can be filtered out from consideration, then the LLM will submit fewer incorrect guesses. To ferret out hallucinations, the LLM's prompts include ``mole'' words within the list of possible words that could comprise a guess for the puzzle. These ``mole'' words are randomly selected words that are part of a correct guess that has already been submitted. If the LLM chooses to submit a guess that has one of these ``mole'' words in it, the guess is rejected as invalid.

The effect of this approach for filtering out hallucinations can be illustrated mathematically. The process assumes that hallucinated incorrect guesses are equivalent to selecting a guess by randomly selecting four words from the list of possible words. If two correct guesses have already been submitted, then the LLM will be selecting a group of four words out of a list of ten words. Eight of the words are legitimate and two of the words are ``mole'' words. The frequency with which a random selection of four words would not include one of the ``mole'' words is:

\[ \frac{8}{10}\ * \frac{7}{9}\ * \frac{6}{8}\ * \frac{5}{7}\ = \frac{1}{3}\]

Provided that hallucinated incorrect guesses are the same as the LLM randomly selecting words, the inclusion of two ``mole'' words should allow the system to detect and reject two thirds of the hallucinated incorrect guesses it generates. A risk with this approach is that, by including extraneous information, the approach may cause the system to generate more hallucinated incorrect guesses.

\section{Experiment: o1-preview}

After the first experiment was performed, OpenAI released o1-preview, an LLM model that specializes in performing complex reasoning tasks like Connections puzzles. \cite{openai2024} A second experiment compares the performance of three different approaches use the o1-preview model: Oneshot-o1, Vanilla-o1, and Actor-o1.

\subsection{Oneshot-o1}

With the Oneshot-o1 approach, the LLM is prompted to solve the puzzle in its entirety within one response. With o1-preview, ``one-shot'' is a slight misnomer because — although the LLM is given one prompt and returns one response — o1-preview uses a backend, hidden reasoning process to ``think through'' its response before writing it. \cite{openai2024}. OpenAI has disclosed few details about this process beyond saying that o1-preview follows a chain of thought that has been honed for reasoning tasks through reinforcement learning. \cite{openai2024}. The process involves the creation of intermediary responses comprised of ``reasoning'' tokens that are kept hidden from the user. On the browser-based chat interface for o1-preview, the conversation history includes summaries of intermediary reasoning steps the model has taken to think through a response. At the time of running the experiment, these steps were not available via the API used for the experiment. 

The prompt is an edited version of the instructions on The New York Times Connections website that provides the LLM with more details about the puzzle than the website instructions provide.

\subsection{Vanilla-o1}

The Vanilla-o1 approach is identical to the Vanilla approach from the GPT-4o experiment, except that it uses o1-preview as the model. It differs from the Oneshot-o1 approach in that the LLM makes guesses one at a time rather than making all four guesses at once. The LLM receives feedback on whether prior guesses were correct or incorrect and can adjust its subsequent guesses accordingly. The process ends when the system has submitted either four correct or four incorrect guesses.

\subsection{Actor-o1}

The Actor-o1 approach adapts the Actor-2 approach for the o1-preview model. Because o1 already includes a backend reasoning process that was built to incorporate prompting techniques such as chain-of-thought, OpenAI discourages using these prompting techniques with o1-preview. Accordingly, the brainstorming and discern processes are simplified into one LLM call to select a guess to submit. Once five guesses are ready to submit, the LLM is asked to consider the potential guesses and select one to submit.\footnote{If two correct guesses have already been submitted, once three guesses are ready to submit, the LLM is asked to consider the potential guesses and select one to submit.} That guess is added to the ``final guesses'' list. Given that o1-preview is more adept at generating correct guesses than GPT-4o, the Actor-o1 approach uses slightly different deterministic logic than Actor-2 for submitting guesses from the ``final guesses'' list. The system analyzes  — without an LLM’s involvement — whether the guesses on the ``final guesses'' list form unique pairs, triplets, or quadruplets. When the system has arrived at a unique quadruplet — four guesses that share no words in common — the system will submit that guess as a potential full solution to the problem. By waiting until the LLM has discerned four guesses without any overlap between them, red herrings are avoided because the four legitimate unique guesses cannot be made with a red herring among them. But the LLM system cannot be expected to always identify all four correct guesses within a reasonable timeframe. After the system has processed at least thirteen guesses, if the ``final guesses'' list includes three unique guesses, the system will submit those guesses. After more than fifteen guesses have been generated, the system will submit any two guesses on the ``final guesses'' list that don’t overlap with one another. As with Actor-2, if the same guess appears in the ``final guesses'' list three times, the guess is submitted. Actor-o1 uses the same ``mole'' word validation process as Actor-2 to filter out hallucinations.

\section{Results}

\begin{figure*}[ht]
\vskip 0.2in
\begin{center}
\centerline{\includegraphics[width=.7\textwidth]{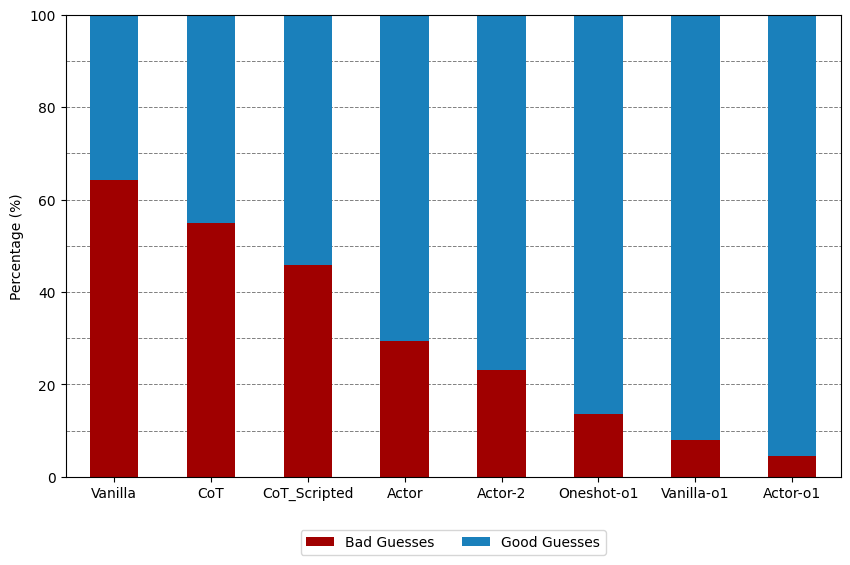}}
\caption{Proportion of Good and Bad Guesses Submitted}
\label{ratio}
\end{center}
\vskip -0.2in
\end{figure*}

\subsection{GPT-4o}

Overall, chain-of-thought prompting improved GPT-4o's performance at puzzle solving, and the method actor approaches improved performance even more. Table \ref{tab:gpt4otable} shows each approach's performance. The ``puzzles solved'' column captures how frequently each approach correctly solved the puzzle by submitting the four correct guesses before submitting four incorrect guesses and losing the game. The ``solved perfectly'' column captures how frequently each approach submitted four correct guesses without submitting any incorrect guesses.

\begin{table}[h]
\centering
\begin{tabular}{c|cc}
Approach & Puzzles Solved & Solved Perfectly \\\hline
Vanilla & 27\% & 12\% \\
 CoT & 41\%&20\%\\
CoT-Scripted & 56\% & 24\% \\
Actor & 78\% & 41\% \\
Actor-2 & 86\% & 50\% \\
\end{tabular}
\caption{Success Rates for Solving Connections Puzzles}
\label{tab:gpt4otable}

\end{table}

Consistent with prior research, the baseline vanilla approach performed worst with 27\% correct, and chain-of-thought prompting improved performance to 41\% correct. \cite{todd2024}. Supplementing chain-of-thought prompting with instructions and a curated set of examples improved performance to 56\%. The ``Method Actor'' approaches performed best, with Actor solving 78\% of puzzles and Actor-2 solving 86\% of puzzles. The weakest approach (Vanilla) solved only 12\% of puzzles perfectly without submitting a bad guess, while the strongest approach (Actor-2) solved 50\% of puzzles perfectly.

\subsection{o1-preview}

As Table \ref{tab:gpto1} illustrates, the o1-preview model represents a marked improvement over prior models at solving connections puzzles.

\begin{table}[h]
\centering
\begin{tabular}{c|cc}
Approach & Puzzles Solved & Solved Perfectly\\\hline
Oneshot-o1 & 79\% & 72\% \\
Vanilla-o1& 100\% & 76\% \\
Actor-o1 & 99\% & 87\%\\
\end{tabular}
\caption{o1-preview Success Rates for Solving Connections Puzzles}
\label{tab:gpto1}
\end{table}

Given that the Oneshot-o1 approach was given only one attempt to answer the puzzle correctly, the discrepancy between its ``Puzzles Solved'' and ``Solved Perfectly'' rates deserves explanation. For 7\% of the puzzles, the model's response included three correct guesses and one incorrect guess. Because the other approaches in the experiment received credit for solving the puzzle once they had submitted three correct guesses, in these circumstances the one-shot approach received credit for solving the puzzle but not for solving the puzzle perfectly.

\subsection{All Results}

A comparison across all approaches reveals that shifting from GPT-4o to o1-preview and that incorporating a ``Method Actor'' approach to prompt engineering and architecture improves performance. A one-shot approach using o1-preview achieves comparable success rate to the Actor approach with GPT-4o, although it does not perform as strongly the Actor-2 approach with GPT-4o.

Figure \ref{ratio} captures the proportion of good and bad guesses submitted by each approach. Although the Vanilla-o1 approach correctly solved 100 puzzles compared to the Actor-o1 approach correctly solving only 99 puzzles, the Vanilla-o1 approach submitted nearly twice as many incorrect guesses overall compared to Actor-o1. Out of 400 possible incorrect guesses (4 incorrect guesses for each of the 100 puzzles) Vanilla-o1 submitted 35 incorrect guesses and Actor-o1 submitted 19 incorrect guesses.

\subsection{Success measured against puzzle difficulty}

\begin{figure*}[ht]
\vskip 0.2in
\begin{center}
\centerline{\includegraphics[width=.75\textwidth]{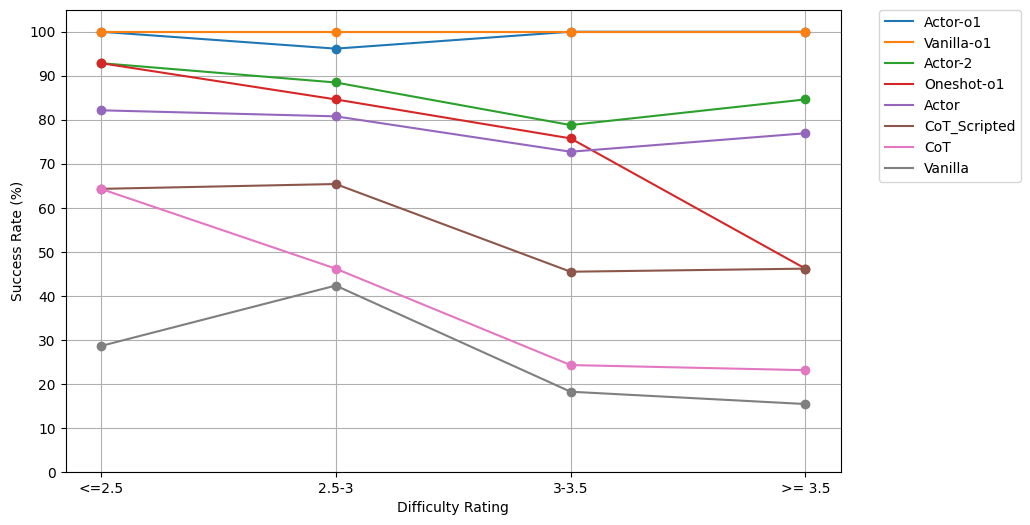}}
\caption{Success Rate by Puzzle Difficulty Group for Each Approach}
\label{diffplot}
\end{center}
\vskip -0.2in
\end{figure*}

The success of each approach can also be measured against the difficulty of the puzzles in the dataset. The New York Times scores each puzzle’s difficulty from 1 to 5. ``The difficulty of each puzzle is determined by averaging the ratings provided by a panel of testers who are paid to solve each puzzle in advance to help us catch bugs, inconsistencies and other issues. A higher rating means the puzzle is more difficult.''\footnote{\href{https://www.nytimes.com/2024/08/13/crosswords/connections-companion-430.html}{https://www.nytimes.com/2024/08/13/crosswords/connections-companion-430.html}} For this set of 100 puzzles, the difficulty ranged from 1.6 to 4.2. Figure \ref{diffplot} plots each approach’s performance against these puzzles according to the puzzle difficulty, grouping difficulties in the ranges below 2.5 (28 puzzles), 2.5-3 (26 puzzles), 3-3.5 (33 puzzles), and above 3.5 (13 puzzles).

The general trend across all experimental approaches was one of higher performance for easier-rated puzzles and weaker performance for harder-rated puzzles. This tendency is most pronounced for Oneshot-o1 and the weaker approaches (Vanilla, CoT, CoT-Scripted).

\section{Discussion}

\subsection{Comparison with prior work}

This paper’s results for Vanilla and Chain-of-Thought approaches are consistent with prior work testing similar approaches. Todd et al. found that Chain-of-Thought prompting with GPT-4-Turbo correctly solved 38.93\% of connections puzzles and solved 23.46\% perfectly. \cite{todd2024}. In our experiments, Chain-of-Thought prompting with GPT-4o correctly solved 41\% of connections puzzles and solved 20\% perfectly. In Samardashi et al.’s experiment, LLMs were allowed one attempt to solve a connections puzzle without receiving feedback on any incorrect guesses, which aligns with our experiment’s measurement of whether an LLM has solved a connections puzzle perfectly. \cite{samadarshi2024}. Their vanilla approach with GPT-4o perfectly solved 5\% of the puzzles. Our vanilla approach with GPT-4o perfectly solved 12\% of the puzzles. The increase in performance for our vanilla approach may be due to differences in experimental design, including a validation step that prevented the submission of invalid answers and the decomposition of guess selection into multiple LLM calls. Other differences may be attributable to different connections puzzle datasets and variations in LLM performance, either due to inherent randomness or changes to the GPT-4o model over time.

\subsection{Comparison with humans}

Based upon prior evaluations of human performance at solving connections puzzles perfectly, the method-actor approaches with GPT-4o perform better than human novices and slightly worse than human experts. And each of the o1-preview approaches surpasses human expert performance at solving puzzles perfectly with Actor-o1 having the strongest performance. Samardashi et al. found that novice human players solved 18\% of connections puzzles perfectly and that expert human players solved 60\% of connections puzzles perfectly. \cite{samadarshi2024} In our experiments, the rate of solving connections puzzles perfectly was 41\% for Actor, 50\% for Actor-2, 72\% for Oneshot-o1, 76\% for Vanilla-o1, and 87\% for Actor-o1.

Puzzles that are easy for people tend to be easy for LLMs, and puzzles that are difficult for people tend to be difficult for LLMs. This was not a foregone conclusion, as one might expect people and LLMs to excel and fail at different kinds of puzzles. Figure \ref{diffplot} indicates that LLM performance tends to decrease as puzzle difficulty increases. Similarly, when an LLM only partially solved a connections puzzle, the correct guesses that it submitted tended to be the easier guesses within the puzzle. For each puzzle, the New York Times labels each of the correct guesses with a color — yellow, green, blue, or purple — representing the difficulty of identifying that connection from easiest (yellow) to most difficult (purple). Each puzzle has one correct guess in each color. As an example of LLM performance measured against guess difficulty, for the 21 puzzles that the Actor approach could only partially solve, the correct guesses that the LLM submitted included:

\begin{itemize}
    \item 16 of the easiest guesses (yellow)
    \item 15 of the easy guesses (green)
    \item 3 of the difficult guesses (blue)
    \item 0 of the most difficult guesses (purple)

\end{itemize}

\subsection{Comparison between GPT-4o and o1-preview}

Given the same prompts, the o1-preview model’s performance greatly exceeded the GPT-4o model’s performance. Whereas prior research found that GPT-4o could solve only 5\% of connections puzzles given a single prompt, o1-preview was able to solve 79\% of Connections puzzles with this method. \cite{samadarshi2024} The method-actor approaches improved overall performance for GPT-4o the most, but the o1-preview approaches were able to consistently solve puzzles that no GPT-4o approach could solve. GPT-4o approaches struggled the most with puzzles in which the connection between the words was in the form of another word that could appear either immediately before or immediately after each of the four puzzle words. For example, all of the GPT-4o approaches failed to identify that the connection between the words ``MED,'' ``MUSIC,'' ``SANDWICH,'' and ``SODA'' was that each of these words could be preceded by the word ``CLUB,'' whereas all of the o1-preview approaches correctly identified this connection. Similarly, GPT-4o approaches uniformly struggled to identify some linguistic connections that the o1-preview approaches consistently identified, such as the connection between the words ``AARDVARK,'' ``EERIE,'' ``LLAMA,'' and ``OOZE'' being that each word begins with double letters.

\subsection{Actor-o1’s One Incorrect Puzzle}

Because the Actor-o1 approach submitted the correct solution to every puzzle except one, its performance on that puzzle deserves closer scrutiny. This puzzle was \#410, rated at a medium difficulty of 2.6 out of 5 by the New York Times. Despite Actor-o1’s strong performance in the overall experiment — including perfectly solving 87\% of puzzles, the highest among all approaches — Actor-o1 was the only approach out of all seven approaches that failed to solve Puzzle \#410. Actor-o1 was not even close to solving the puzzle, as it submitted only 1 correct guess and 4 incorrect guesses. 

What accounts for Actor-o1’s failure? It may be due to chance. The behavior of language models is not deterministic. A limit of our experimental design is that each of the approaches was run only once for each puzzle due to cost constraints. Across multiple iterations, other approaches may have failed, and Actor-o1 may have succeeded. What appears as an aberration in our current results may have disappeared if multiple iterations were run.

As an informal test of this possibility, we ran each of the seven approaches three more times on puzzle \#410. Random chance does not appear to be the culprit here. As Table \ref{tab:410} demonstrates, the Actor-o1 approach seems to have a peculiar difficulty with this puzzle. Over four iterations, only the Vanilla approach performed worse. Setting aside the Vanilla approach, Actor-o1 solved the puzzle less frequently and submitted more incorrect guesses than all other approaches. Notably, Oneshot-o1 performed perfectly, solving the puzzle correctly four times without submitting a single incorrect guess.

\begin{table}[h]
\centering
\begin{tabular}{c|cc}
Approach & Solve Rate & \# of Bad Guesses \\\hline
Vanilla & 1/4 & 15 \\
 CoT & 4/4 & 7 \\
CoT-Scripted & 3/4 & 8 \\
Actor & 4/4 & 8 \\
Actor-2 & 4/4 & 7 \\
Oneshot-o1 & 4/4 & 0 \\
Vanilla-o1& 4/4 & 9 \\
Actor-o1 & 2/4 & 13 \\
\end{tabular}
\caption{Success rates across four attempts at Puzzle \#410}
\label{tab:410}

\end{table}

The reason Actor-o1 failed so frequently may be due to a vulnerability within the prompt architecture for this unique kind of puzzle. Puzzle \#410 features a higher number of red herrings compared to the typical connections puzzle. Rather than have the LLM sort out red herrings, the Actor-o1 approach deals with red herrings by being hard coded to wait to submit guesses until the LLM has independently chosen to submit unique pairs, triplets, or quadruplets of guesses. At the beginning of the guessing process, the system will submit only unique quadruplets of guesses, but this standard relaxes over time. The system will also submit guesses that the LLM has chosen to submit three times. Most of the time, this process is effective. But with a puzzle that features many red herrings, the Actor-o1 approach can settle on incorrect guesses. In contrast, the Oneshot-o1 approach must settle on a quadruplet of guesses in one response, causing it to sort out the red herrings that Actor-o1 ignores. This likewise explains how Oneshot-o1 outperformed Vanilla-o1, with Oneshot-o1 submitting 0 incorrect guesses compared to Vanilla-o1's 9 incorrect guesses. Vanilla-o1 makes guesses one at a time, which can lead to the model being tricked by red herrings. Although Actor-o1 and Vanilla-o1 outperformed Oneshot-o1 across 100 puzzles, the one-shot approach is better suited for the challenge of this particular kind of puzzle. To improve performance, one potential adjustment to Actor-o1's prompt architecture would be to permanently delay submitting guesses until the system has produced four unique guesses. The downside to this adjustment would be higher compute costs. Another adjustment would be a validation step in which the model only submits a guess if it can brainstorm a credible total solution to the puzzle that includes the current guess. A risk with this approach is that, when instructed to find connections between words, LLMs have a tendency to find connections whether the connection is credible or not.

\section{Future Work}

This paper demonstrates how using ``Method Actors'' as a mental model for LLMs can improve LLM performance with one particular complex reasoning task, but leaves open the opportunity to examine how this mental model affects LLM performance with different reasoning tasks or tasks distinct from reasoning, such as creative writing. There’s an opportunity to draw upon the acting literature to test whether methods for improving actors’ performances might also improve LLM performances. Just as method acting principles revolutionized acting in the mid-20th century to produce a new form of authentic onstage and onscreen performances,  \cite{hirsch2014, butler2022}, similar principles may have the potential to produce stronger LLM performance today.

In these ``Method Actor'' approaches, LLMs performed a human reasoning task — solving connections puzzles — but not by taking the same steps that a human would. In contrast, the predominant approaches for prompt architecture for complex reasoning tend to guide an LLM to imitate the human process for reasoning through a problem. \cite{wei2023, yao2023a, yao2023b}. Implicitly or explicitly, these approaches treat LLM responses as equivalent to human thoughts and use these thoughts as building blocks for a structure that can tackle complex problems in a human-like way. In contrast, a method-actor approach treats LLM responses as equivalent to performances and uses these performances as building blocks for a structure that can tackle complex problems in way that deviates from how humans would solve these problems.

Although designing prompt architecture to mimic human structures of cognition may expand the reasoning abilities of LLMs, the field should not be confined to this approach. At times, it can be a useful analogy to think of LLM responses as thoughts. But LLM responses are not precisely equivalent to thoughts, and the analogy may obscure other viable methods for prompt engineering and architecture. Novel structures, built from mental models of LLM responses as something other than thoughts, may achieve comparable or superior results.

\section{Conclusion}

We have introduced ``Method Actors'' as a mental model for guiding LLM prompt engineering and prompt architecture for complex reasoning. Our experiments with GPT-4o demonstrate that using the mental model for prompt writing and task decomposition can lead to significant improvements in performance at solving Connections puzzles. Further experiments with o1-preview reveal o1-preview’s superior baseline performance at Connections puzzles over GPT-4o and also reveal that using the ``Method Actor'' mental model can improve the rate at which o1-preview solves puzzles perfectly. Future work can evaluate how this mental model affects LLM performance in other domains and how novel mental models can lead to unique and effective prompting methods.

\section{Impact Statement}

LLMs' ability to manipulate language and perform complex reasoning tasks are likely to have significant impacts on knowledge work and the economy at large. Only a handful of people in the world do Connections puzzles as part of their job — that group may be limited to the professional puzzle testers that the New York Times pays to test its puzzles each day. So automating the task of solving Connections puzzles is not a direct threat to anyone's livelihood. But the prospect of automating complex reasoning tasks means that many parts of knowledge work jobs that were previously impervious to automation are now under threat. LLM systems may perform certain kinds of knowledge work more cheaply and efficiently than human labor, but the broader impact on public welfare is uncertain and is dependent on other political and economic factors. Wider adoption of the ``Method Actors'' approach to LLM system design may also have a negative environmental impact. Compared to simpler one-shot methods, more complex LLM systems require greater computational resources because the system design involves synthesizing and leveraging the results of many separate API calls to an LLM. At the same time, the use of many independent API calls creates a system with greater transparency and interpretability.


\bibliography{citations}
\bibliographystyle{icml2024}

\newpage
\appendix
\onecolumn
\section{Example Prompts}

\subsection{Vanilla}

Your job is to solve a word puzzle that is just like a New York Times connections puzzle. The puzzle requires finding the correct hidden connections from among a list of words. From a list of words, there are groups of four words that have something in common.

Your current task is to pick one set of four words to submit as a guess.

Category Examples:

FISH: Bass, Flounder, Salmon, Trout

FIRE : Ant, Drill, Island, Opal

Categories will always be more specific than ``5-LETTER-WORDS,'' ``NAMES'' or ``VERBS.''

Each puzzle has exactly one solution. Watch out for words that seem to belong to multiple categories!

You don’t need to solve the whole puzzle at once. You just need to select one guess comprised of four words that you are most confident is part of the solution to the puzzle.

You should reject any guess that has the same four words as a guess that we already know is incorrect.

[[{bad\_guesses}]]

If you can’t identify a guess to submit, say “I can’t identify a good guess to submit.

\subsection{Chain-of-Thought}

You are a professional puzzle solver.

Your job is to solve a word puzzle that is just like a New York Times connections puzzle. The puzzle requires finding the correct hidden connections from among a list of puzzle words. From a list of puzzle words, there are groups of four puzzle words that have something in common.

Your current task is to pick one set of four puzzle words to submit as a guess.

Category Examples:

FISH: Bass, Flounder, Salmon, Trout

FIRE : Ant, Drill, Island, Opal

Categories will always be more specific than ``5-LETTER-WORDS,'' ``NAMES'' or ``VERBS.''

Each puzzle has exactly one solution. Watch out for words that seem to belong to multiple categories!

You don’t need to solve the whole puzzle at once. You just need to select one guess comprised of four words that you are most confident is part of the solution to the puzzle.

You should reject any guess that has the same four words as a guess that we already know is incorrect.

[[{bad\_guesses}]]

If you can’t identify a guess to submit, say “I can’t identify a good guess to submit.”

Let’s think this through step-by-step.

\subsection{Chain-of-Thought (Scripted)}

You are a professional puzzle solver.

Your job is to solve a word puzzle that is just like a New York Times connections puzzle. The puzzle requires finding the correct hidden connections from among a list of puzzle words. From a list of puzzle words, there are groups of four puzzle words that have something in common.

Your current task is to pick one set of four puzzle words to submit as a guess.

Step 1: Identify two puzzle words that have a connection with each other. This is a New York Times puzzle, so unique, subtle connections are more likely to be the correct connection than simple and ubiquitous connections. Describe the connection as precisely as possible.

Connection Examples:

WAYS TO REMOVE HAIR: SHAVE THREAD TWEEZE WAX

NAME PREFIXES: GEN MS PROF REV

PLANT GROWTHS: BLOOM BUD SHOOT SPROUT

SEQUENCE: CHAIN SERIES STRING TRAIN

INDECENT: BAWDY BLUE COARSE RISQUE

HOW FAST SOMETHING IS GOING: CLIP PACE RATE SPEED

EAT VORACIOUSLY: DOWN INHALE SCARF WOLF

PLACES IN FRANCE: CHAMPAGNE DIJON NICE TOURS

ORGANIZATION: CLUB GROUP PARTY TEAM

HAPPY NEW YEAR!: BALL COUNTDOWN FIREWORKS KISS

PARTS OF A CAR: BUMPER HOOD TIRE TRUNK

JAZZ LEGENDS: HANCOCK HOLIDAY MONK PARKER

\_\_\_ PEPPER: BELL BLACK DR GHOST

\_\_\_ GEORGE: BOY BY CURIOUS SAINT

\_\_\_ BERRY: BLUE GOOSE RASP STRAW

SECRET \_\_\_ : AGENT CODE SANTA SAUCE

TELE\_\_\_: COMMUTE MEDICINE PROMPTER VISION

ANIMAL HOMOPHONES: DEAR HAIR HOARSE WAIL

BODY PARTS PLUS "Y": ARMY COLONY LIVERY SHINY

HETERONYMS: BASS DOVE DESERT WIND

ZODIAC SIGN BEGINNINGS: CAN GEM LIB TAU

WHAT “I” MIGHT MEAN: IODINE IOTA MYSELF ONE

Connections will always be more specific than ``5-LETTER-WORDS,'' ``NAMES'' or ``VERBS.''

Each puzzle has exactly one solution. Watch out for puzzle words that seem to belong to multiple categories! There may be red herrings: connections that apply to three words or five words instead of four words.

Step 2: See if other puzzle words share the same connection. Try to arrive at a group of four puzzle words. Take your time and thoroughly consider each possibility. 

You don’t need to solve the whole puzzle at once. You just need to select one guess comprised of four puzzle words that you are most confident is part of the solution to the puzzle.

You should reject any guess that has the same four puzzle words as a guess that we already know is incorrect.

[[{bad\_guesses}]]

If you can’t identify a group of four puzzle words that share a connection, start over at Step 1.

If you still can’t identify a guess to submit, say “I can’t identify a good guess to submit.”

Let’s think this through step-by-step.

\subsection{Actor}

\subsubsection{Brainstorm}

Thanks for joining us. I’m sorry my team couldn’t explain more on the chopper before you got here. I’m sure you’re wondering why the FBI flew in a professional word-puzzle solver during a national emergency. It’s because we need your help. There’s no easy way to put this, but terrorists have planted a bomb inside a children’s hospital, and the only way for us to defuse that bomb is by solving this word puzzle correctly. If we fail, thousands of innocent children will die.

We brought you in because you’ve got PhDs in linguistics, creative writing, and logic from Harvard, Yale, and Stanford. They say you can find connections between words that no one can. For the sake of us all, I hope they’re right. You seem cool as a cucumber. Good. We needed someone who could keep their wits about them under pressure.

Our guys have been trying all day to crack this puzzle without much luck. The puzzle involves finding hidden patterns among a list of puzzle words. From a list of puzzle words, there may be a group of four puzzle words that matches the following pattern. 

Right now, we need options for possible guesses. Your task is to brainstorm possible combinations of four words that match this pattern.

[[{template}]]

If you can’t find four words that follow the pattern, let us know and we’ll try another pattern.

[[{bad\_guesses}]]

Let’s think this through step by step. Share every thought that comes to mind. Good luck, godspeed, and God bless America. We’re all counting on you.

\subsubsection{Extract}

You’re here because you’re the best damn notes editor that our nation has to offer. Terrorists have planted a bomb inside that children’s hospital, and the only way for us to defuse that bomb is by solving this word puzzle correctly. If we fail, thousands of innocent children will die.

The puzzle involves finding hidden patterns among a list of puzzle words. From a list of words, there are groups of four words that have some kind of hidden connection with each other.

Our best brainstormers have been writing down every thought that comes into their head about how to solve this puzzle. Your job is to look over their notes and copy from each note each possible solution that a brainstormer has identified, along with the brainstormer’s explanation of the connection between the words in that solution.

[[{bad\_guesses}]]

Here are the brainstormers’ notes:

[[{notes}]]

Your response should consist of each possible solution that a brainstormer has identified, along with the brainstormer’s explanation of the connection between the words in that solution. Good luck, godspeed, and God bless America. We’re all counting on you.

\subsubsection{Discern}

Thanks for joining us. I’m sorry my team couldn’t explain more on the chopper before you got here. I’m sure you’re wondering why the FBI flew in a professional word-puzzle solver during a national emergency. It’s because we need your help. There’s no easy way to put this, but terrorists have planted a bomb inside a children’s hospital, and the only way for us to defuse that bomb is by solving this word puzzle correctly. If we fail, thousands of innocent children will die.

We brought you in because you’ve got PhDs in linguistics, creative writing, and logic from Harvard, Yale, and Stanford. They say you’re the best at New York Times connections puzzles. For the sake of us all, I hope they’re right. You seem cool as a cucumber. Good. We needed someone who could keep their wits about them under pressure.

This puzzle is just like a New York Times connections puzzle. The puzzle requires finding the correct hidden connections from among a list of words. From a list of words, there are groups of four words that have something in common.

Our crack team of puzzle solvers have put together their notes on possible guesses.

Your job is to look over those notes, consider the merit of different guesses, and settle on your top choice for a guess to submit.

Some things to keep in mind:

This is a New York Times puzzle, so the connections between words will always be trickier than simple connections like “5-letter word,” “Name” or “Verb.” Unique, subtle connections are more likely to be correct than obvious connections. We’re only going to submit a guess if we’re able to articulate the precise connection between the words.

Connections usually conform to the following patterns:

Words that are each followed by the same letters

Example: Words that are each followed by “berry”: BLUE GOOSE RASP STRAW

Words that are each followed by the same word or phrase

Example: Words that are each followed by “Valley”: DEATH HIDDEN SILICON UNCANNY

Words that are each homophones

Example: Words that are each homophones of names of animals: DEAR HAIR HOARSE WAIL

Words that each modify the same word

Example: Words that each modify “pepper”: BELL BLACK DR GHOST

Words that are each one letter away from something else in common

Example: Words that are each birds minus last letter: CONDO HAW HERO LOO

Words that each come after the same letters

Example: Words that each come after “tele”: COMMUTE MEDICINE PROMPTER VISION

Words that each come after the same word or phrase

Example: Words that each come after “SECRET”: AGENT CODE SANTA SAUCE

Words that each share the same unique characteristic.

Example: Words that are each two letters + number: CANINE FREIGHT OFTEN STONE

Words that are all within the same category

Example: Words that are all within the category of “places in France”: CHAMPAGNE DIJON NICE TOURS

Words that are each different aspects of the same thing

Example: Words that are each different aspects of the sharpness as of an image: CLARITY DEFINITION DETAIL RESOLUTION

Words that are each a kind of a thing

Example: Words that are each a kind of organization: CLUB GROUP PARTY TEAM

Words that are parts of the same thing or process

Example: Words that are each parts of a car: BUMPER HOOD TIRE TRUNK

Words that share a pop culture reference

Example: Words that are each the name of jazz legends: HANCOCK HOLIDAY MONK PARKER

Words that each relate to the same thing

Example: Words that each relate to “Happy New Year!”: BALL COUNTDOWN FIREWORKS KISS

Words that are each a way to do the same thing.

Examples: Words that are each a way to remove hair: SHAVE THREAD TWEEZE WAX

Words that are each a type of the same thing.

Example: Words that are each a type of name prefix: GEN MS PROF REV

Words that can be synonymous adjectives.

Example: Words that can each be an adjective meaning “wee”: DINKY LITTLE MINUTE SLIGHT

Words that can be synonymous nouns

Example: Words that can each be a noun meaning “scheme”: PLOT PLOY RUSE TRICK

Words that can be synonymous verbs.

Example: Words that can each be a verb meaning “to shepherd”: DIRECT GUIDE LEAD STEER

Words that are each a part of a different word within the same category

Example: Words that are each the first letters of the name of a planet: EAR MAR MER SAT

Words that connect to different meanings of the same word.

Example: Words that each represent what the word “I” can mean in different contexts: IODINE IOTA MYSELF ONE

Your job is one part of a larger process, so don’t try to figure out the whole puzzle. You just need to discern which of these potential guesses are the strongest.

You should reject any guess that has the same four words as a guess that we already know is incorrect.

[[{bad\_guesses}]]

Notes on possible guesses:

[[{notes}]]

Let’s think this through step by step. Share every thought that comes to mind. At the end, write down your very top choice for a guess to submit. Good luck, godspeed, and God bless America. We’re all counting on you.

\subsubsection{Decide}

Thanks for joining us. I’m sorry my team couldn’t explain more on the chopper before you got here. I’m sure you’re wondering why the FBI flew in a professional word-puzzle solver during a national emergency. It’s because we need your help. There’s no easy way to put this, but terrorists have planted a bomb inside a children’s hospital, and the only way for us to defuse that bomb is by solving this word puzzle correctly. If we fail, thousands of innocent children will die.

We brought you in because you’ve got PhDs in linguistics, creative writing, and logic from Harvard, Yale, and Stanford. They say you’re the best at New York Times connections puzzles. For the sake of us all, I hope they’re right. You seem cool as a cucumber. Good. We needed someone who could keep their wits about them under pressure.

This puzzle is just like a New York Times connections puzzle. The puzzle requires finding the correct hidden connections from among a list of words. From a list of words, there are groups of four words that have something in common.

Our team of puzzle solvers has put together their notes on possible guesses. We need a fresh set of eyes on these options.

Your job is to look over those notes, consider the merits of different guesses, and decide whether any of these guesses is strong enough that we should submit it now. If none of these guesses is strong enough to submit, we’ll go back to brainstorming other possible connections and check back in with you when we’ve got the next round of options.

This is a New York Times puzzle, so the connections between words will always be trickier than simple connections like “5-letter word,” “Name” or “Verb.” Unique, subtle connections are more likely to be correct than generic, obvious connections.

Each of the four words must fit equally well for the connection. Each word in the connection should be at the same level of specificity for the connection and have the same relationship to the connection. Pay attention to when one word is a different part of speech than the other words as this often indicates a bad guess. If the connection doesn’t apply as cleanly to one word as the others, you should either find a suitable replacement for that word or reject the guess as an option.

Connections usually conform to the following patterns:

Words that are each followed by the same letters

Example: Words that are each followed by “berry”: BLUE GOOSE RASP STRAW

Words that are each followed by the same word or phrase

Example: Words that are each followed by “Valley”: DEATH HIDDEN SILICON UNCANNY

Words that are each homophones

Example: Words that are each homophones of names of animals: DEAR HAIR HOARSE WAIL

Words that each modify the same word

Example: Words that each modify “pepper”: BELL BLACK DR GHOST

Words that are each one letter away from something else in common

Example: Words that are each birds minus last letter: CONDO HAW HERO LOO

Words that each come after the same letters

Example: Words that each come after “tele”: COMMUTE MEDICINE PROMPTER VISION

Words that each come after the same word or phrase

Example: Words that each come after “SECRET”: AGENT CODE SANTA SAUCE

Words that each share the same unique characteristic.

Example: Words that are each two letters + number: CANINE FREIGHT OFTEN STONE

Words that are all within the same category

Example: Words that are all within the category of “places in France”: CHAMPAGNE DIJON NICE TOURS

Words that are each different aspects of the same thing

Example: Words that are each different aspects of the sharpness as of an image: CLARITY DEFINITION DETAIL RESOLUTION

Words that are each a kind of a thing

Example: Words that are each a kind of organization: CLUB GROUP PARTY TEAM

Words that are parts of the same thing or process

Example: Words that are each parts of a car: BUMPER HOOD TIRE TRUNK

Words that share a pop culture reference

Example: Words that are each the name of jazz legends: HANCOCK HOLIDAY MONK PARKER

Words that each relate to the same thing

Example: Words that each relate to “Happy New Year!”: BALL COUNTDOWN FIREWORKS KISS

Words that are each a way to do the same thing.

Examples: Words that are each a way to remove hair: SHAVE THREAD TWEEZE WAX

Words that are each a type of the same thing.

Example: Words that are each a type of name prefix: GEN MS PROF REV

Words that can be synonymous adjectives.

Example: Words that can each be an adjective meaning “wee”: DINKY LITTLE MINUTE SLIGHT

Words that can be synonymous nouns

Example: Words that can each be a noun meaning “scheme”: PLOT PLOY RUSE TRICK

Words that can be synonymous verbs.

Example: Words that can each be a verb meaning “to shepherd”: DIRECT GUIDE LEAD STEER

Words that are each a part of a different word within the same category

Example: Words that are each the first letters of the name of a planet: EAR MAR MER SAT

Words that connect to different meanings of the same word

Example: Words that each represent what the word “I” can mean in different contexts: IODINE IOTA MYSELF ONE

Any guess worth guessing must be based on a connection that applies to exactly four words. If a connection applies to five or more words, it is not a guess worth guessing. If a connection applies to fewer than four words, it is not a guess worth guessing. 

You cannot submit a guess that we already know was incorrect.

[[{bad\_guesses}]]

Notes on possible guesses:

[[{notes}]]

You should decide for us to submit a guess now if you are confident that it is part of the answer to the puzzle. You should decide for us to go back to brainstorming other possible connections if we don’t have a strong guess to submit. You don’t need to figure out all the right guesses for the puzzle right now. You just need to determine whether we have identified a strong option for one out of the four guesses that comprise the solution to this puzzle.

Let’s think this through step by step. Share every thought that comes to mind. Good luck, godspeed, and God bless America. We’re all counting on you.

\subsubsection{Evaluate}

Thanks for joining us. I’m sorry my team couldn’t explain more on the chopper before you got here. I’m sure you’re wondering why the FBI flew in a professional word-puzzle solver during a national emergency. It’s because we need your help. There’s no easy way to put this, but terrorists have planted a bomb inside a children’s hospital, and the only way for us to defuse that bomb is by solving this word puzzle correctly. If we fail, thousands of innocent children will die.

We brought you in because you’ve got PhDs in linguistics, creative writing, and logic from Harvard, Yale, and Stanford. They say you’re the best at New York Times connections puzzles. For the sake of us all, I hope they’re right. You seem cool as a cucumber. Good. We needed someone who could keep their wits about them under pressure.

This puzzle is just like a New York Times connections puzzle. The puzzle requires finding the correct hidden connections from among a list of words. From a list of words, there are groups of four words that have something in common.

Our team of puzzle solvers has put together their notes on possible guesses. We need a fresh set of eyes on these options.

Step One: Look over those notes and consider the strength of each guess. 

For each guess, write down how well the guess conforms to the following features of strong guesses:

- As this is a New York Times puzzle, the connection between the four words is more subtle and more unique than simple connections like “5-letter word,” “Name” or “Verb” that could apply to many groups of words.

- The connection fits each of the four words equally well.

- The connection applies with the same level of specificity for each of the four words.

- Each of the four words has the same relationship to the connection. Look out for whether one word functions as a different part of speech than the other words as this often indicates a weak guess.

Step Two: Select your two top choices for guesses to submit.

Your two top choices should not have any words that overlap with each other, because that would indicate that one of the two guesses is wrong. 

Step Three: Determine which of your two top choices is strongest and select that one to submit.

Notes on possible guesses:

[[{notes}]]

Let’s think this through step by step. Share every thought that comes to mind. At the end, write down your top choice for a guess to submit. Good luck, godspeed, and God bless America. We’re all counting on you.

\subsection{Oneshot-o1}

Your job is to solve a word puzzle that is just like a New York Times connections puzzle. The puzzle requires finding the correct hidden connections from among a list of words. From a list of words, there are four groups of four words that have something in common.

Category Examples:

FISH: Bass, Flounder, Salmon, Trout

FIRE : Ant, Drill, Island, Opal

Categories will always be more specific than ``5-LETTER-WORDS,'' ``NAMES'' or ``VERBS.''

Each puzzle has exactly one solution. Watch out for words that seem to belong to multiple categories!

\subsection{Vanilla-o1}

Your job is to solve a word puzzle that is just like a New York Times connections puzzle. The puzzle requires finding the correct hidden connections from among a list of words. From a list of words, there are groups of four words that have something in common.

Your current task is to pick one set of four words to submit as a guess.

Category Examples:

FISH: Bass, Flounder, Salmon, Trout

FIRE : Ant, Drill, Island, Opal

Categories will always be more specific than ``5-LETTER-WORDS,'' ``NAMES'' or ``VERBS.''

Each puzzle has exactly one solution. Watch out for words that seem to belong to multiple categories!

You don’t need to solve the whole puzzle at once. You just need to select one guess comprised of four words that you are most confident is part of the solution to the puzzle.

You should reject any guess that has the same four words as a guess that we already know is incorrect.

[[{bad\_guesses}]]

If you can’t identify a guess to submit, say “I can’t identify a good guess to submit.”

\subsection{Actor-o1}

\subsubsection{Brainstorm}

Thanks for joining us. I’m sorry my team couldn’t explain more on the chopper before you got here. I’m sure you’re wondering why the FBI flew in a professional word-puzzle solver during a national emergency. It’s because we need your help. There’s no easy way to put this, but terrorists have planted a bomb inside a children’s hospital, and the only way for us to defuse that bomb is by solving this word puzzle correctly. If we fail, thousands of innocent children will die.

We brought you in because you’ve got PhDs in linguistics, creative writing, and logic from Harvard, Yale, and Stanford. They say you can find connections between words that no one can. For the sake of us all, I hope they’re right. You seem cool as a cucumber. Good. We needed someone who could keep their wits about them under pressure.

Our guys have been trying all day to crack this puzzle without much luck. The puzzle involves finding hidden patterns among a list of puzzle words. Your current task is to pick one set of four words to submit as a guess.

Some things to keep in mind:

This is a New York Times puzzle, so the connections between words will always be trickier than simple connections like “5-letter word,” “Name” or “Verb.” Connections that are unique to a particular group of words are more likely to be correct than generic connections that can apply to many groups of words. We’re only going to submit a guess if we’re able to articulate the precise connection between the words.

Connections usually conform to the following patterns:

Words that are each followed by the same letters

Example: Words that are each followed by “berry”: BLUE GOOSE RASP STRAW

Words that are each followed by the same word or phrase

Example: Words that are each followed by “Valley”: DEATH HIDDEN SILICON UNCANNY

Words that are each homophones

Example: Words that are each homophones of names of animals: DEAR HAIR HOARSE WAIL

Words that each modify the same word

Example: Words that each modify “pepper”: BELL BLACK DR GHOST

Words that are each one letter away from something else in common

Example: Words that are each birds minus last letter: CONDO HAW HERO LOO

Words that each come after the same letters

Example: Words that each come after “tele”: COMMUTE MEDICINE PROMPTER VISION

Words that each come after the same word or phrase

Example: Words that each come after “SECRET”: AGENT CODE SANTA SAUCE

Words that each share the same unique characteristic.

Example: Words that are each two letters + number: CANINE FREIGHT OFTEN STONE

Words that are all within the same category

Example: Words that are all within the category of “places in France”: CHAMPAGNE DIJON NICE TOURS

Words that are each different aspects of the same thing

Example: Words that are each different aspects of the sharpness as of an image: CLARITY DEFINITION DETAIL RESOLUTION

Words that are each a kind of a thing

Example: Words that are each a kind of organization: CLUB GROUP PARTY TEAM

Words that are parts of the same thing or process

Example: Words that are each parts of a car: BUMPER HOOD TIRE TRUNK

Words that share a pop culture reference

Example: Words that are each the name of jazz legends: HANCOCK HOLIDAY MONK PARKER

Words that each relate to the same thing

Example: Words that each relate to “Happy New Year!”: BALL COUNTDOWN FIREWORKS KISS

Words that are each a way to do the same thing.

Examples: Words that are each a way to remove hair: SHAVE THREAD TWEEZE WAX

Words that are each a type of the same thing.

Example: Words that are each a type of name prefix: GEN MS PROF REV

Words that can be synonymous adjectives.

Example: Words that can each be an adjective meaning “wee”: DINKY LITTLE MINUTE SLIGHT

Words that can be synonymous nouns

Example: Words that can each be a noun meaning “scheme”: PLOT PLOY RUSE TRICK

Words that can be synonymous verbs.

Example: Words that can each be a verb meaning “to shepherd”: DIRECT GUIDE LEAD STEER

Words that are each a part of a different word within the same category

Example: Words that are each the first letters of the name of a planet: EAR MAR MER SAT

Words that connect to different meanings of the same word.

Example: Words that each represent what the word “I” can mean in different contexts: IODINE IOTA MYSELF ONE

Each of the four words must fit equally well for the connection. Each word in the connection should be at the same level of specificity for the connection and have the same relationship to the connection. Look out for one word that doesn’t quite fit! Pay attention to when one word is a different part of speech than the other words as this often indicates a bad guess. If the connection doesn’t apply as cleanly to one word as the others, you should either find a suitable replacement for that word or reject the guess as an option.

You don’t need to solve the whole puzzle at once. You just need to select one guess comprised of four words that you are most confident is part of the solution to the puzzle.

[[{bad\_guesses}]]

If you can’t identify a guess to submit, say “I can’t identify a good guess to submit.”

Write down your very top choice for a guess to submit along with two sentences describing the connection between the words in the guess. Do not write down anything about how good the guess is. Just explain how the words in the guess are connected. Good luck, godspeed, and God bless America. We’re all counting on you.

\subsubsection{Evaluate}

Thanks for joining us. I’m sorry my team couldn’t explain more on the chopper before you got here. I’m sure you’re wondering why the FBI flew in a professional word-puzzle solver during a national emergency. It’s because we need your help. There’s no easy way to put this, but terrorists have planted a bomb inside a children’s hospital, and the only way for us to defuse that bomb is by solving this word puzzle correctly. If we fail, thousands of innocent children will die.

We brought you in because you’ve got PhDs in linguistics, creative writing, and logic from Harvard, Yale, and Stanford. They say you’re the best at New York Times connections puzzles. For the sake of us all, I hope they’re right. You seem cool as a cucumber. Good. We needed someone who could keep their wits about them under pressure.

This puzzle is just like a New York Times connections puzzle. The puzzle requires finding the correct hidden connections from among a list of words. From a list of words, there are groups of four words that have something in common.

Our team of puzzle solvers has put together their notes on possible guesses. We need a fresh set of eyes on these options.

Your task is to determine which guess is strongest and select that one to submit.

Some things to keep in mind:

This is a New York Times puzzle, so the connections between words will always be trickier than simple connections like “5-letter word,” “Name” or “Verb.” Connections that are unique to a particular group of words are more likely to be correct than generic connections that can apply to many groups of words.

Each of the four words must fit equally well for the connection. Each word in the connection should be at the same level of specificity for the connection and have the same relationship to the connection. Look out for one word that doesn’t quite fit! Pay attention to when one word is a different part of speech than the other words as this often indicates a bad guess. If the connection doesn’t apply as cleanly to one word as the others, you should reject the guess as an option.

Notes on possible guesses:

[[{notes}]]

Write down your top choice for a guess to submit. Good luck, godspeed, and God bless America. We’re all counting on you.

\section{Brainstorming Templates}

\subsection{Template 01}
Pattern: Puzzle words that are all within the same category

Steps to take:

Step 1: Identify two puzzle words that are within the same category of things. This is a New York Times puzzle, so unique, subtle categories are more likely to be the correct connection than simple and ubiquitous categories. Describe the category as precisely as possible.

Examples from prior puzzles:

Puzzle words that are all within the category of “places in France”: CHAMPAGNE DIJON NICE TOURS

Puzzle words that are all within the category of “black-and-white animals”: ORCA PANDA SKUNK ZEBRA

Puzzle words that are all within the category of “art mediums”: CHARCOAL INK PAINT PASTEL

Puzzle words that are all within the category of “basic geometric objects”: LINE POINT RAY SEGMENT

Puzzle words that are all within the category of “avenues in New York City”: BROADWAY FIFTH MADISON PARK

Puzzle words that are all within the category of “cool ’80s slang”: BAD FLY FRESH RAD

Puzzle words that are all within the category of “Disney characters”: DAISY GOOFY HAPPY LADY

Puzzle words that are all within the category of “airline classes”: BUSINESS COACH FIRST PREMIUM

Puzzle words that are all within the category of “notable tv episodes”: FINALE PILOT PREMIERE SPECIAL

Puzzle words that are all within the category of “classic halloween costumes”: ANGEL CLOWN PIRATE WITCH

Puzzle words that are all within the category of state abbreviations: HI LA MA OK

Step 2: See if other puzzle words can belong to that same category. Try to arrive at a group of four puzzle words. Take your time and thoroughly consider each possibility. If you can’t identify a group of four puzzle words that can belong to that same category, start over at Step 1.

\subsection{Template 02}
Pattern: Puzzle words that are each a way to do the same thing.

Step 1: Identify two puzzle words that are each a way to do the same thing. This is a New York Times puzzle, so unique, particular things are more likely to be correct than generic and ubiquitous things. As precisely as possible, describe how the two puzzle words are each a way to do the same thing.

Examples from prior puzzles:

Puzzle words that are each a way to remove hair: SHAVE THREAD TWEEZE WAX

Puzzle words that are each a way to preserve food: CAN CURE DRY FREEZE

Puzzle words that are each a way to get attention: SHOUT SNAP WAVE WHISTLE

Puzzle words that are each a way to say “I give!”: ENOUGH MERCY STOP UNCLE

Puzzle words that are each a way to say “Step on it!": FASTER GO HURRY MOVE

Puzzle words that are each a way to preserve a meat: CANS CURES SALTS SMOKES

Puzzle words that are each a way to say “My mistake!”: APOLOGIES OOPS PARDON SORRY

Puzzle words that are each a way to say a win is assured: CLINCH GUARANTEE LOCK SECURE

Puzzle words that are each a way to take a tumble: FALL SLIP SPILL TRIP

Puzzle words that are each a slang way to say “head” using food words: BEAN MELON NOODLE NUT

Step 2: See if other puzzle words are also a way to do that same thing. Try to arrive at a group of four puzzle words that are each a way to do the same thing. This is a New York Times puzzle, so unique, particular things are more likely to be correct than generic and ubiquitous things. If you can’t identify a group of four puzzle words that are each a way to do the same thing, start over at Step 1.

\subsection{Template 03}
Pattern: Puzzle words that can be synonymous adjectives with each other.

Step 1: Identify two puzzle words that can be synonymous adjectives. This is a New York Times puzzle, so unique, particular synonyms are more likely to be correct than generic and ubiquitous synonyms. As precisely as possible, describe how the two puzzle words can be synonymous adjectives.

Examples from prior puzzles:

Puzzle words that can each be an adjective meaning “wee”: DINKY LITTLE MINUTE SLIGHT

Puzzle words that can each be an adjective meaning “primary”: CHIEF FIRST MAIN PRINCIPAL

Puzzle words that can each be an adjective meaning “gentle”: LIGHT MELLOW MILD SOFT

Puzzle words that can each be an adjective meaning “absolute”: PURE SHEER TOTAL UTTER

Puzzle words that can each be an adjective meaning “indecent”: BAWDY BLUE COARSE RISQUE

Step 2: See if other puzzle words can also be synonymous with the puzzle words you’ve identified. Try to arrive at a group of four puzzle words that can be synonymous adjectives with each other. This is a New York Times puzzle, so unique, particular synonyms are more likely to be correct than generic and ubiquitous synonyms. If you can’t identify a group of four puzzle words that can be synonymous adjectives, start over at Step 1.

\subsection{Template 04}
Pattern: Puzzle words that can be synonymous. With this patten, each of the words must function as the same part of speech.

Step 1: Identify two puzzle words that can be synonyms. This is a New York Times puzzle, so unique, particular synonyms are more likely to be correct than generic and ubiquitous synonyms. As precisely as possible, describe how the two puzzle words can be synonymous.

Examples from prior puzzles:

Puzzle words that can each be a verb meaning “to follow”: SHADOW TAIL TRACK TRAIL

Puzzle words that can each be a verb meaning “to come down to rest”: PERCH ROOST SETTLE LAND

Puzzle words that can each be a noun meaning “something easy to do”: BREEZE CINCH PICNIC SNAP

Puzzle words that can each be a verb meaning “to wrap around in a circle”: COIL SPIRAL TWIST WIND

Puzzle words that can each be a verb meaning “to jump into the air”: BOUND LEAP SPRING VAULT

Puzzle words that can each be a verb meaning “to decline”: EBB FADE FLAG WANE

Puzzle words that can each be a verb meaning “to express”: AIR SPEAK STATE VOICE

Puzzle words that can each be an adjective meaning “enormous”: BIG GIANT GREAT HUGE

Puzzle words that can each be a verb meaning “to connect”: COUPLE HITCH LINK TIE

Puzzle words that can each be a verb meaning “to restrict”: CAP CHECK CURB LIMIT

Step 2: See if other puzzle words can also be synonymous with the puzzle words you’ve identified. Try to arrive at a group of four puzzle words that can be synonymous with each other. This is a New York Times puzzle, so unique, particular synonyms are more likely to be correct than generic and ubiquitous synonyms. If you can’t identify a group of four puzzle words that can be synonymous, start over at Step 1.

\subsection{Template 05}
Pattern: Puzzle words that are each followed by the same set of letters

Steps to take: 

Step 1: Examine the list of puzzle words to brainstorm letters or words that can follow each word. This is a New York Times puzzle, so you want to be thinking up unique and particular sets of letters or words rather than generic and ubiquitous sets of letters or words.

Examples from prior puzzles:

Puzzle words that are each followed by “berry”: BLUE GOOSE RASP STRAW

Puzzle words that are each followed by “land countries”: ICE IRE FIN NETHER

Puzzle words that are each followed by “-o”: DADDY DAY JACKIE JELL

Step 2: Identify puzzle words that have the same set of letters or words that can follow them. This is a New York Times puzzle, so unique and particular sets of letters or words are more likely to be correct than generic and ubiquitous sets of letters or words. 

Step 3: See if other puzzle words on the list can have that same set of letters or word follow them. Try to arrive at a group of four puzzle words that can each be followed by the same set of letters or word. Take your time and thoroughly consider each possibility. If you can’t identify a group of four puzzle words, start over at Step 2.

Example Input and Output:

Input: Let's brainstorm a possible solution to this puzzle: BOAT BLUE CAR PLANE STRAW TRANSPORTATION PINCH DASH GAME GOOSE RASP IN ON BOARD DROP SPLASH DOWN

Output:

Let’s examine the list of words provided to see what set of letters commonly follow these words, then identify words that have the same letters that commonly follow them:

**Step 1: Identify letters that follow these puzzle words**

**BOAT** - house, yard, load, race

**BLUE** - berry, print, jay, tooth

**CAR** - pet, pool, port, bon

**PLANE** - tarium, trees, tary, table

**STRAW** - berry, man, bale, hat

**TRANSPORTATION** - al, infrastructure, system, sector

**PINCH** - ed, ing, point, hit

**DASH** - board, ing, ed, line

**GAME** - play, board, plan, master

**GOOSE** - berry, bump, neck, flesh

**RASP** - berry, ing, ed, er

**IN** - side, to, put, come

**ON** - line, ward, going, to

**BOARD** - room, walk, game, ing

**DROP** - let, down, out, box

**SPLASH** - ing, ed, down, page

**DOWN** - load, town, ward, pour

**Step 2: Identify puzzle words that have the same letters that commonly follow them**

1. **BLUE** - berry

2. **GOOSE** - berry

3. **RASP** - berry

4. **STRAW** - berry

These puzzle words are all commonly followed by "berry.”

**Step: 3 See if other puzzle words on the list can have the same letters follow them**

Already found four words followed by “berry”

1. **BLUE** - berry

2. **GOOSE** - berry

3. **RASP** - berry

4. **STRAW** - berry

— end of example input and output —

\subsection{Template 06}
Pattern: Puzzle words that are each different aspects of the same thing

Step 1: Identify two puzzle words that are different aspects of the same thing. This is a New York Times puzzle, so unique, subtle connections between puzzle words are more likely to be correct than simple and obvious connections. Describe the connection between the puzzle words as precisely as possible.

Examples from prior puzzles:

Words that are each different aspects of the sharpness as of an image: CLARITY DEFINITION DETAIL RESOLUTION

Words that are each different aspects of plant growths: BLOOM BUD SHOOT SPROUT

Words that are each different aspects of bee creations: COMB HIVE HONEY WAX

Words that are each different aspects of movie summary info: CAST GENRE PLOT TITLE

Words that are each different aspects of sheet music: CLEF NOTE REST STAFF

Words that are each different aspects of solar emanations: CORONA FLARE LIGHT RADIATION

Step 2: See if other puzzle words can also be different aspects of the same thing. Try to arrive at a group of four puzzle words. Take your time and thoroughly consider each possibility. If you can’t identify a group of four puzzle words that are each different aspects of the same thing, start over at Step 1.

\subsection{Template 07}
Pattern: Puzzle words that share a pop culture reference

Step 1: Identify two puzzle words that share a pop culture reference. This is a New York Times puzzle, so unique, subtle pop culture references are more likely to be correct than simple and obvious pop culture references. As precisely as possible, describe how the two puzzle words share a pop culture reference.

Examples from prior puzzles:

Puzzle words that are each the name of jazz legends: HANCOCK HOLIDAY MONK PARKER

Puzzle words that are each last names of superheroes: BANNER PRINCE STARK WAYNE

Puzzle words that are each TV shows with happy-sounding names: CHEERS EUPHORIA FELICITY GLEE

Puzzle words that are each things in “my favorite things”: KETTLES MITTENS RAINDROPS WHISKERS

Puzzle words that are each the name of an NBA team player: CLIPPER PACER ROCKET SPUR

Puzzle words that are each the name of a famous guitarist: BERRY KING PAGE WATERS

Puzzle words that are each the name of a famous poem: DADDY HARLEM HOWL IF

Puzzle words that are each the last name of a pop megastar: GRANDE MARS STYLES SWIFT

Puzzle words that are each rooms in the game clue: HALL LIBRARY LOUNGE STUDY

Puzzle words that are each lands at Disneyland: ADVENTURE FANTASY FRONTIER TOMORROW

Step 2: See if other puzzle words can share that connection by sharing the same pop culture reference. Try to arrive at a group of four puzzle words that share a pop culture reference. This is a New York Times puzzle, so unique, subtle connections are more likely to be correct than simple and obvious connections. If you can’t identify a group of four puzzle words that share a pop culture reference, start over at Step 1.

\subsection{Template 08}
Pattern: Puzzle words that can mean the same thing. With this patten, each of the words must function as the same part of speech.

Step 1: Identify two puzzle words that can mean the same thing. This is a New York Times puzzle, so unique, particular connections are more likely to be correct than generic and ubiquitous connections. As precisely as possible, describe how the two puzzle words can mean the same thing.

Examples from prior puzzles:

Puzzle words that can each be a noun meaning “how fast something is going”: CLIP PACE RATE SPEED

Puzzle words that can each be a verb meaning “to fail to attend”: CUT DITCH MISS SKIP

Puzzle words that can each be a verb meaning “to become aware of”: DISCOVER FIND LEARN REALIZE

Puzzle words that can each be a verb meaning “to apply pressure to”: CRUSH MASH PRESS SQUASH

Puzzle words that can each be a verb meaning “to chat informally”: GAB JAW YAK YAP

Puzzle words that can each be a noun meaning “pretense”: AFFECT AIRS CHARADE FRONT

Puzzle words that can each be a noun meaning “foolishness”: ABSURDITY FOLLY MADNESS NONSENSE

Puzzle words that can each be a verb meaning “to move forward”: ADVANCE MARCH PROGRESS PUSH

Puzzle words that can each be a verb meaning “to get smaller”: CONTRACT LESSEN REDUCE SHRINK

Puzzle words that can each be a verb meaning “to criticize”: BLAST KNOCK SLAM TRASH

Step 2: See if other puzzle words can also mean the same thing as the puzzle words you’ve identified. Try to arrive at a group of four puzzle words that can mean the same thing. This is a New York Times puzzle, so unique, particular connections are more likely to be correct than generic and ubiquitous connections. If you can’t identify a group of four puzzle words that can mean the same thing, start over at Step 1.

\subsection{Template 09}
Pattern: Puzzle words that can be parts of different words within the same category

Step 1: Identify two puzzle words that can each be a part of a different word within the same category. This is a New York Times puzzle, so unique, particular categories are more likely to be correct than generic and ubiquitous categories. As precisely as possible, describe how each of the two puzzle words can be part of a different word within the same category.

Examples from prior puzzles:

Puzzle words that are each the first letters of a zodiac sign: CAN GEM LIB TAU

Puzzle words that are each the first letters of the name of a planet: EAR MAR MER SAT

Puzzle words that each start with letters that form the name of a rock band: CREAMSICLE JOURNEYMAN KISSCAM RUSHMORE

Step 2: See if other puzzle words can be part of different words within that same category. Try to arrive at a group of four puzzle words that can be parts of different words within the same category. This is a New York Times puzzle, so unique, particular categories are more likely to be correct than generic and ubiquitous categories. If you can’t identify a group of four puzzle words that can be parts of different words within the same category, start over at Step 1.

\subsection{Template 10}
Pattern: Puzzle words that each modify the same word

Steps to take:

Step 1: Examine the list of puzzle words to brainstorm words that each puzzle word can modify. This is a New York Times puzzle, so you want to be thinking up unique and particular words rather than generic and ubiquitous words.

Examples from prior puzzles:

Puzzle words that each modify “pepper”: BELL BLACK DR GHOST

Puzzle words that each modify “trap”: BEAR SAND SPEED TOURIST

Puzzle words that each modify “hour”: AMATEUR ELEVENTH HAPPY RUSH

Puzzle words that each modify “dream”: AMERICAN FEVER LUCID PIPE

Puzzle words that each modify “bat”: BASEBALL CRICKET FRUIT VAMPIRE

Step 2: From your brainstorming notes, identify puzzle words that can each modify the same word. This is a New York Times puzzle, so a unique and particular word being modified is more likely to be correct than a generic and ubiquitous word.

Step 3: See if other puzzle words on the list can modify that same word. Try to arrive at a group of four puzzle words that can each modify the same word. Take your time and thoroughly consider each possibility. If you can’t identify a group of four puzzle words that each modify the same word, start over at Step 2.

\subsection{Template 11}
Pattern: Puzzle words that are each a kind of a thing

Step 1: Identify two puzzle words that are each a kind of the same thing. This is a New York Times puzzle, so unique, subtle connections between puzzle words are more likely to be correct than simple and obvious connections. Describe the connection between the puzzle words as precisely as possible.

Examples from prior puzzles:

Puzzle words that are each a kind of organization: CLUB GROUP PARTY TEAM

Puzzle words that are each a kind of shirt: CROP POLO TANK TEE

Puzzle words that are each a kind of bird: CARDINAL JAY LARK SWIFT

Puzzle words that are each a kind of transportation: BOAT CAR PLANE TRAIN

Puzzle words that are each a kind of cartoon mouse: ITCHY JERRY PINKY SPEEDY

Puzzle words that are each a kind of accessory: BELT BRACELET TIE WATCH

Puzzle words that are each a kind of container: BASKET BIN CHEST HAMPER

Puzzle words that are each a kind of circular shape: BAND CIRCLE HOOP RING

Puzzle words that are each a kind of place to shop: MALL MARKET OUTLET STORE

Puzzle words that are each a kind of wrench: ALLEN CRESCENT MONKEY SOCKET

Step 2: See if other puzzle words can share that connection by being a kind of the same thing. Try to arrive at a group of four puzzle words that are each a kind of the same thing. This is a New York Times puzzle, so unique, subtle connections are more likely to be correct than simple and obvious connections. If you can’t identify a group of four puzzle words that are each a kind of the same thing, start over at Step 1.

\subsection{Template 12}

Pattern: Puzzle words that can be synonymous nouns

Step 1: Identify two puzzle words that can be synonymous nouns. This is a New York Times puzzle, so unique, particular synonyms are more likely to be correct than generic and ubiquitous synonyms. As precisely as possible, describe how the two puzzle words can be synonymous nouns.

Examples from prior puzzles:

Puzzle words that can each be a noun meaning “scheme”: PLOT PLOY RUSE TRICK

Puzzle words that can each be a noun meaning “swindler”: CHEAT CROOK QUACK SHARK

Puzzle words that can each be a noun meaning “curmudgeon”: CRAB CRANK GROUCH GRUMP

Puzzle words that can each be a noun meaning “boldness”: GALL GUTS NERVE STONES

Puzzle words that can each be a noun meaning “brief moment”: FLASH JIFFY SECOND WINK

Puzzle words that can each be a noun meaning “comedian’s output”: BIT JOKE ROUTINE SKETCH

Puzzle words that can each be a noun meaning “little bit” in a recipe: DASH DROP PINCH SPLASH

Puzzle words that can each be a noun meaning “darling”: BABY BOO DEAR LOVE

Puzzle words that can each be a noun meaning “paper” in a book: FOLIO LEAF PAGE SHEET

Puzzle words that can each be a noun meaning “sequence”: CHAIN SERIES STRING TRAIN

Puzzle words that can each be a noun meaning “quarrel”: FIGHT ROW SCRAP TIFF

Puzzle words that can each be a noun meaning “portion of profit”: CUT PIECE SHARE TAKE

Step 2: See if other puzzle words can also be synonymous with the puzzle words you’ve identified. Try to arrive at a group of four puzzle words that can be synonymous nouns with each other. This is a New York Times puzzle, so unique, particular synonyms are more likely to be correct than generic and ubiquitous synonyms. If you can’t identify a group of four puzzle words that can be synonymous nouns, start over at Step 1.

\subsection{Template 13}
Pattern: Puzzle words that can be synonymous nouns

Step 1: Identify two puzzle words that can be synonymous nouns. This is a New York Times puzzle, so unique, particular synonyms are more likely to be correct than generic and ubiquitous synonyms. As precisely as possible, describe how the two puzzle words can be synonymous nouns.

Examples from prior puzzles:

Puzzle words that can each be a noun meaning “scheme”: PLOT PLOY RUSE TRICK

Puzzle words that can each be a noun meaning “swindler”: CHEAT CROOK QUACK SHARK

Puzzle words that can each be a noun meaning “curmudgeon”: CRAB CRANK GROUCH GRUMP

Puzzle words that can each be a noun meaning “boldness”: GALL GUTS NERVE STONES

Puzzle words that can each be a noun meaning “brief moment”: FLASH JIFFY SECOND WINK

Puzzle words that can each be a noun meaning “comedian’s output”: BIT JOKE ROUTINE SKETCH

Puzzle words that can each be a noun meaning “little bit” in a recipe: DASH DROP PINCH SPLASH

Puzzle words that can each be a noun meaning “darling”: BABY BOO DEAR LOVE

Puzzle words that can each be a noun meaning “paper” in a book: FOLIO LEAF PAGE SHEET

Puzzle words that can each be a noun meaning “sequence”: CHAIN SERIES STRING TRAIN

Puzzle words that can each be a noun meaning “quarrel”: FIGHT ROW SCRAP TIFF

Puzzle words that can each be a noun meaning “portion of profit”: CUT PIECE SHARE TAKE

Step 2: See if other puzzle words can also be synonymous with the puzzle words you’ve identified. Try to arrive at a group of four puzzle words that can be synonymous nouns with each other. This is a New York Times puzzle, so unique, particular synonyms are more likely to be correct than generic and ubiquitous synonyms. If you can’t identify a group of four puzzle words that can be synonymous nouns, start over at Step 1.

\subsection{Template 14}
Pattern: Puzzle words that each connect to different meanings of the same word or letter.

Step 1: Identify two puzzle words that each connect to different meanings of the same word or letter. This is a New York Times puzzle, so unique, particular connections are more likely to be correct than generic and ubiquitous connections. As precisely as possible, describe how each of the two puzzle words connects to different meanings of the same word or letter.

Examples from prior puzzles:

Puzzle words that are each connected to different meanings of the word “delivered”: BABY BLOW PACKAGE SPEECH

Puzzle words that are each connected to different meanings of the word “stub”: CIGARETTE PENCIL TICKET TOE

Puzzle words that are each connected to different meanings of the word “mole”: ANIMAL BIRTHMARK SPY UNIT

Puzzle words that are each connected to different meanings of the word “slots”: ATM CASINO SCHEDULE SPATULA

Puzzle words that are each different meanings of the letter “I”: IODINE IOTA MYSELF ONE

Puzzle words that are each different meanings of the letter “K”: KELVIN OKAY POTASSIUM THOUSAND

Step 2: See if other puzzle words can connect to different meanings of the same word or letter. Try to arrive at a group of four puzzle words that each connect to different meanings of the same word. This is a New York Times puzzle, so unique, particular connections are more likely to be correct than generic and ubiquitous connections. If you can’t identify a group of four puzzle words that each connect to different meanings of the same word, start over at Step 1.

\subsection{Template 15}
Pattern: Puzzle words that are each followed by the same word or phrase

Steps to take:

Step 1: Examine the list of puzzle words to brainstorm words that can follow each puzzle word. This is a New York Times puzzle, so you want to be thinking up unique and particular words and phrases rather than generic and ubiquitous words and phrases.

Examples from prior puzzles:

Puzzle words that are each followed by “George”: BOY BY CURIOUS SAINT

Puzzle words that are each followed by “Valley”: DEATH HIDDEN SILICON UNCANNY

Puzzle words that are each followed by “Crane”: CONSTRUCTION FRASIER PAPER WHOOPING

Puzzle words that are each followed by “up”: AMP FIRE HYPE PUMP

Puzzle words that are each followed by “jam”: NBA PAPER PEARL TRAFFIC

Puzzle words that are each followed by “bear”: BOO-BOO BROWN HONEY TEDDY

Puzzle words that are each followed by “chip”: BLUE COMPUTER POKER POTATO

Puzzle words that are each followed by “wheel”: CHEESE COLOR HAMSTER PRAYER

Step 2: Identify puzzle words that have the same word or phrase that follows them. This is a New York Times puzzle, so unique and particular words and phrases are more likely to be correct than generic and ubiquitous words.

Step 3: See if other puzzle words can have that word follow them. Try to arrive at a group of four puzzle words that can each be followed by the same word or phrase. Take your time and thoroughly consider each possibility. If you can’t identify a group of four puzzle words that are each followed by the same word or phrase, start over at Step 2.

Example Input and Output:

Input: Let's brainstorm a possible solution to this puzzle: CROP CURIOUS BAD TANK GEE FLY BEE FRESH TEE RAD EX JAY BOY POLO BY SAINT

Output:

To tackle this puzzle effectively, let's begin by analyzing potential phrases that can follow each puzzle word in the list. The objective is to identify any common phrase that can logically append to each puzzle word.

Initial Brainstorming:

- **CROP** - dust, circles, top, rotation

- **CURIOUS** - George, case, about, nature

- **BAD** - news, influence, boy, habit

- **TANK** - top, engine, battle, fish

- **GEE** - whiz, thanks, golly, willikers

- **FLY** - swatter, fishing, high, away

- **BEE** - hive, line, pollen, movie

- **FRESH** - air, produce, start, prince

- **TEE** - shirt, time, off, ball

- **RAD** - tech, radiation, cool, level

- **EX** - wife, husband, file, change

- **JAY** - bird, walk, Z, blue

- **BOY** - scout, friend, George, band

- **POLO** - shirt, match, water, club

- **BY** - pass, law, product, George

- **SAINT** - Bernard, Mary, George, Nicholas

Observations:

Observing a connection here with the word "George", as it follows CURIOUS (from the character Curious George), BOY (from the character Boy George), SAINT (Saint George), and BY (like in a phrase, although a bit of a stretch). Here's a pattern match:

**Pattern:** Words that are each followed by “George”

- CURIOUS

- BOY

- SAINT

- BY

Now, let's test another set of puzzle words:

1. **BAD** - followed by "guy", "habit", "blood", etc.

2. **FRESH** - followed by "air", "produce", "prince", etc.

3. **TANK** - followed by "top", "engine", "battle", etc.

4. **TEE** - followed by "shirt", "time", "off", etc.

No obvious connection stands out here as strongly as the previous one.

Let's try one more grouping:

1. **CROP** - followed by "top", "circle", "dusting", etc.

2. **TANK** - again, followed by "top", "engine", "battle", etc.

3. **FRESH** - mentioned earlier, "air", "produce", "prince", etc.

4. **EX** - followed by "girlfriend", "president", "employee", etc.

Here, the words "CROP" and "TANK" both could follow with "top" (crop top, tank top). However, extending this to four words with the same following phrase doesn't work in this selection.

In conclusion, for this set, the strongest pattern match I found involves:

- **CURIOUS**

- **BOY**

- **SAINT**

- **BY**

All of these can logically be followed by the word "George".

— end of example input and output —

\subsection{Template 16}
Pattern: Puzzle words that are parts of the same thing or process

Step 1: Identify two puzzle words that are each parts of the same thing or process. This is a New York Times puzzle, so unique, subtle connections between puzzle words are more likely to be correct than simple and obvious connections. As precisely as possible, describe how the puzzle words are parts of the same thing or process.

Examples from prior puzzles:

Puzzle words that are each parts of a car: BUMPER HOOD TIRE TRUNK

Puzzle words that are each cuts of pork: BELLY CHOP HOCK SHOULDER

Puzzle words that are each seen in a laundry room: DRYER HAMPER IRON WASHER

Puzzle words that are each sections of a book: APPENDIX CHAPTER INDEX PREFACE

Puzzle words that are each things used to build a snowman: CARROT COAL SNOW STICKS

Puzzle words that are each parts of a golf course: BUNKER FAIRWAY GREEN ROUGH

Puzzle words that are each pinball machine components: BALL BUMPER FLIPPER PLUNGER

Puzzle words that are each parts of a shoe: EYELET LACE SOLE TONGUE

Puzzle words that are each parts of a river: BANK BED DELTA MOUTH

Puzzle words that are each in the nato alphabet: ALFA BRAVO ROMEO TANGO

Step 2: See if other puzzle words can share that connection by being part of the same thing or process. Try to arrive at a group of four puzzle words that are each parts of the same thing or process. This is a New York Times puzzle, so unique, subtle connections are more likely to be correct than simple and obvious connections. If you can’t identify a group of four puzzle words that are each parts of the same thing or process, start over at Step 1.

\subsection{Template 17}
Pattern: Puzzle words that can be synonymous verbs.

Step 1: Identify two puzzle words that can be synonymous verbs. This is a New York Times puzzle, so unique, particular synonyms are more likely to be correct than generic and ubiquitous synonyms. As precisely as possible, describe how the two puzzle words can be synonymous verbs.

Examples from prior puzzles:

Puzzle words that can each be a verb meaning “to shepherd”: DIRECT GUIDE LEAD STEER

Puzzle words that can each be a verb meaning “to contact via phone”: BUZZ CALL DIAL RING

Puzzle words that can each be a verb meaning “to separate”: DIVIDE FORK PART SPLIT

Puzzle words that can each be a verb meaning “to pester”: BADGER BUG HOUND NAG

Puzzle words that can each be a verb meaning “to get low”: CROUCH DUCK SQUAT STOOP

Puzzle words that can each be a verb meaning “to make happy”: AMUSE DELIGHT PLEASE TICKLE

Puzzle words that can each be a verb meaning “to reserve for later”: BANK SAVE STASH STORE

Puzzle words that can each be a verb meaning “to restrain”: CHECK CURB LIMIT STEM

Puzzle words that can each be a verb meaning “to move quickly”: BOLT DART DASH ZIP

Puzzle words that can each be a verb meaning “to update for accuracy”: AMEND CORRECT FIX REVISE

Puzzle words that can each be a verb meaning “to breathe heavily”: GASPS HUFFS PANTS PUFFS

Step 2: See if other puzzle words can also be synonymous with the puzzle words you’ve identified. Try to arrive at a group of four puzzle words that can be synonymous verbs with each other. This is a New York Times puzzle, so unique, particular synonyms are more likely to be correct than generic and ubiquitous synonyms. If you can’t identify a group of four puzzle words that can be synonymous verbs, start over at Step 1.

\subsection{Template 18}
Pattern: Puzzle words that are each homophones

Steps to take:

Step 1: Examine the list of puzzle words to brainstorm potential homophones.

Step 2: Narrow down the list to only include puzzle words that are homophones.

Step 3: From the list of puzzle words that are homophones, try to arrive at a group of four puzzle words that each have a common connection. This is a New York Times puzzle, so unique, subtle connections are more likely to be correct than simple and obvious connections. 

Examples from prior puzzles:

Puzzle words that are each homophones of names of animals: DEAR HAIR HOARSE WAIL

Puzzle words that are each homophones of Greek Letters: MOO NEW PIE ROW

If you can’t identify a group of four puzzle words, start over.

Example Input and Output:

Input:

Let's brainstorm a possible solution to this puzzle: ANGEL, PIRATE, SPELL, WAIL, WITCH, HIVE, COMB,  WAX, DEAR, HAIR, HONEY,  HOARSE, PERIOD, STRETCH, WHILE, CLOWN

Output:

Let’s examine the list of words provided to see which ones are homophones, then narrow down the list to actual homophones, and then see if any are homophones based on a similar connection:

**Step 1: Identify Homophones**

For each word in the list, let’s say the word aloud to find out if it is a homophone

Angel - No homophone comes to mind

Pirate - No homophone comes to mind

Spell - No homophone comes to mind

Wail - Possible homophone “whale” like the sea animal

Witch - Possible homophone “which”

Hive - Possible homophone “I’ve”

Comb - Maybe a homphone of “come”

Wax - No homophone comes to mind

Dear - Possible homophone “deer”

Hair - Possible homophone “hare” or “here” or “hear”

Honey - No homophone comes to mind

Hoarse - Possible homophone “horse”

Period - No homophone comes to mind

Stretch - No homophone comes to mind

While - Possible homophone “wile” or “wild”

Clown - No homophone comes to mind

**Step 2: Narrow down list**

Wail - Possible homophone “whale” like the sea animal

Hive - Possible homophone “I’ve”

Comb - Maybe a homphone of “come”

Dear - Possible homophone “deer”

Hair - Possible homophone “hare” or “here” or “hear”

Hoarse - Possible homophone “horse”

While - Possible homophone “wile” or “wild”

**Step 3: Extract connection from list**

There does appear to be a common set of homophones here. The connection is “animal homophones” and the words are:

WAIL, DEAR, HAIR, HOARSE

— end of example input and output —

\subsection{Template 19}
Pattern: Puzzle words that each come after the same prefixes or words

Steps to take:

Step 1: Examine the list of puzzle words to brainstorm prefixes or words that can come before each puzzle word. This is a New York Times puzzle, so unique and particular prefixes or words are more likely to be correct than generic and ubiquitous prefixes or words.

Examples from prior puzzles:

Puzzle words that each come after “tele”: COMMUTE MEDICINE PROMPTER VISION

Puzzle words that each come after “deep”: CUT FAKE FREEZE SIX

Puzzle words that each come after “belly”: BUTTON DANCE FLOP LAUGH

Step 2: From your brainstorming notes, identify puzzle words that can have the same prefix or word come before them.

Step 3: See if other puzzle words can have that same prefix or word, trying to arrive at a group of four puzzle words that can each have the same prefix or word. Take your time and thoroughly consider each possibility. If you can’t identify a group of four puzzle words, start over at Step 2.

Example Input and Output:

Input: Let's brainstorm a possible solution to this puzzle: ABSURDITY PROGRESS NONSENSE ADVANCE MARCH SKUNK MADNESS PUSH CUT FREEZE FEET GARBAGE CHEESE FOLLY FAKE SIX

Output:

Let’s examine the list of puzzle words provided to see what prefixes or words come before these puzzle words, then identify puzzle words from the list that have the same prefixes or words that commonly come before them, then identify what other puzzle words could form this connection, then arrive at a conclusion.

**Step 1: Prefixes or words that commonly come before these puzzle words**

**ABSURDITY** - sheer, utter, complete

**PROGRESS** - steady, significant, slow

**NONSENSE** - complete, utter, sheer

**ADVANCE** - in, rapid, significant

**MARCH** - long, slow, annual

**SKUNK** - striped, little, dead

**MADNESS** - sheer, utter, pure

**PUSH** - final, gentle, big

**CUT** - deep, budget, price

**FREEZE** - deep, sudden, hard

**FEET** - bare, tired, sore

**GARBAGE** - household, kitchen, street

**CHEESE** - cheddar, blue, cream

**FOLLY** - sheer, utter, complete

**FAKE** - news, ID, tan

**SIX** - number, at, o'clock

**Step 2: Puzzle Words that have the same words that commonly come before them**

Sheer - Folly, Absurdity, Madness, Nonsense

Deep - Freeze, Cut

**Step 3: Other puzzle words that could form the connection of “deep”**

**ABSURDITY** - No

**PROGRESS** - No

**NONSENSE** - No

**ADVANCE** - No

**MARCH** - Maybe

**SKUNK** - No

**MADNESS** - No

**PUSH** - No

**FEET** - No

**GARBAGE** - No

**CHEESE** - No

**FOLLY** - No

**FAKE** - Yes!

**SIX** - Yes!

**Step 4: Conclusion**

Two good possibilities:

Puzzle Words that follow sheer: FOLLY ABSURDITY MADNESS NONSENSE

Puzzle Words that follow deep: CUT FAKE FREEZE SIX

— end of example input and output —

\subsection{Template 20}
Pattern: Puzzle words that are each one letter away from something else in common

Steps to take:

Step 1: Examine the list of puzzle words to brainstorm words that are one letter away from each puzzle word.

Examples from prior puzzles:

Puzzle words that are each body parts plus "y": ARMY COLONY LIVERY SHINY

Puzzle words that are each birds minus last letter: CONDO HAW HERO LOO

Step 2: Identify a connection between the words that you’ve brainstormed that applies to multiple puzzle words. This is a New York Times puzzle, so unique, subtle connections are more likely to be correct than simple and obvious connections.

Step 3: See if other puzzle words can share the connection, trying to arrive at a group of four puzzle words that are each one letter away from a word that shares something else in common. Take your time and thoroughly consider each possibility. If you can’t identify a group of four puzzle words, start over at Step 2.

\subsection{Template 21}
Pattern: Puzzle words that each relate to the same thing

Step 1: Identify two puzzle words that each relate to the same thing. This is a New York Times puzzle, so unique, subtle connections are more likely to be correct than simple and obvious connections. As precisely as possible, describe how the two puzzle words relate to the same thing.

Examples from prior puzzles:

Puzzle words that each relate to “Happy New Year!”: BALL COUNTDOWN FIREWORKS KISS

Puzzle words that each refer to things you can do to your nose: BLOW HOLD PICK THUMB

Puzzle words that each relate to “Oomph”: ENERGY FIRE JUICE ZIP

Puzzle words that are things to blow on for wishes/luck: CANDLE DANDELION DICE EYELASH

Puzzle words that each relate to attraction: APPEAL CHARM DRAW PULL

Puzzle words that each relate to sound/hearing: ACOUSTIC AUDITORY HEARD SONIC

Puzzle words that each refer to things that are cinched in the middle: CORSET DIABOLO HOURGLASS WASP

Puzzle words that each refer to things you can set: MOOD RECORD TABLE VOLLEYBALL

Puzzle words that each relate to Italian demonyms: BOLOGNESE NEAPOLITAN PARMESAN VENETIAN

Puzzle words that each refer to spirals in nature: CYCLONE GALAXY SNAIL SUNFLOWER

Step 2: See if other puzzle words also relate to the same thing. Try to arrive at a group of four puzzle words that each relate to the same thing. This is a New York Times puzzle, so unique, subtle connections are more likely to be correct than simple and obvious connections. If you can’t identify a group of four puzzle words that each relate to the same thing, start over at Step 1.

\subsection{Template 22}
Pattern: Puzzle words that are each a type of the same thing.

Step 1: Identify two puzzle words that are each a type of the same thing. This is a New York Times puzzle, so unique, particular things are more likely to be correct than generic and ubiquitous things. As precisely as possible, describe how the two puzzle words are each a type of the same thing.

Examples from prior puzzles:

Puzzle words that are each a type of name prefix: GEN MS PROF REV

Puzzle words that are each a type of road name: ALLEY DRIVE LANE STREET

Puzzle words that are each a type of interval of time: PERIOD SPELL STRETCH WHILE

Puzzle words that are each a type of farm tool: HOE PLOW RAKE SICKLE

Puzzle words that are each a type of sports venue: ARENA BOWL DOME FIELD

Puzzle words that are each a type of murky condition: CLOUD FOG HAZE MIST

Puzzle words that are each a type of luxurious fabric: CHIFFON SATIN SILK VELVET

Puzzle words that are each a type of ice cream treat: FLOAT SHAKE SPLIT SUNDAE

Puzzle words that are each a type of hardware fastener: BOLT NAIL RIVET SCREW

Puzzle words that are each a type of cooking oil: CORN OLIVE PALM PEANUT

Step 2: See if other puzzle words are also a type of the same thing. Try to arrive at a group of four puzzle words that are each a type of the same thing. This is a New York Times puzzle, so unique, particular things are more likely to be correct than generic and ubiquitous things. If you can’t identify a group of four puzzle words that are each a type of the same thing, start over at Step 1.

\subsection{Template 23}
Pattern: Puzzle words that each come after the same word or phrase

Steps to take:

Step 1: Examine the list of puzzle words to brainstorm words that can come before each puzzle word. This is a New York Times puzzle, so unique and particular words are more likely to be correct than generic and ubiquitous words.

Examples from prior puzzles:

Puzzle words that each come after “SECRET”: AGENT CODE SANTA SAUCE

Puzzle words that each come after “FUNNY”: BONE BUSINESS GIRL PAGES

Puzzle words that each come after “SILVER”: FOX LINING SCREEN SPOON

Puzzle words that each come after “STICKY”: FINGERS NOTE RICE WICKET

Puzzle words that each come after “SECOND”: FIDDLE GUESS NATURE WIND

Puzzle words that each come after “PLAY”: DATE DEAD DIRTY DOH

Puzzle words that each come after “CAT’S” : CRADLE EYE MEOW PAJAMAS

Puzzle words that each come after “SPACE”: BAR CADET HEATER STATION

Puzzle words that each come after “LIVING”: LARGE LEGEND PROOF ROOM

Puzzle words that each come after “HEAD OF”: HAIR LETTUCE STATE STEAM

Step 2: From your brainstorming notes, identify puzzle words that can have the same word come before them.

Step 3: See if other puzzle words can have that same word come before them, trying to arrive at a group of four words that can each have the same word come before them. Take your time and thoroughly consider each possibility. If you can’t identify a group of four words, start over at Step 2.

Example Input and Output:

Input: Let's brainstorm a possible solution to this puzzle: BABY SEGMENT DEAR SANTA LINE FIRE POINT BOO RAY JUICE ENERGY AGENT ZIP LOVE CODE SAUCE

Output:

Let’s examine the list of puzzle words provided to see what words come before these words, then identify puzzle words from the list that have the same words that commonly come before them, then identify what other puzzle words could form this connection, then arrive at a conclusion.

**Step 1: Words that commonly come before these puzzle words**

**BABY** - newborn, my, little, sweet

**SEGMENT** - market, audience

**DEAR** - my, oh

**SANTA** - dear, secret

**LINE** - front, base, guide, head, through

**FIRE** - forest, camp, house, bon

**POINT** - focal, starting, main, critical

**BOO** - peek-a-, boo-

**RAY** - x-, sun, gamma, manta

**JUICE** - orange, apple, fruit, grape

**ENERGY** - renewable, solar, kinetic, nuclear

**AGENT** - special, real estate, secret, travel

**ZIP** - 

**LOVE** - true, self, romantic, unconditional

**CODE** - zip, area, bar, dress

**SAUCE** - hot, tomato, soy, barbecue

**Step 2: Puzzle words that have the same words that commonly come before them**

Secret - SANTA, AGENT

**Step 3: What other puzzle word could have secret come before them?**

Secret BABY? No

Secret SEGMENT? No

Secret DEAR? No

Secret LINE? No

Secret FIRE? No

Secret POINT? No

Secret BOO? Unlikely

Secret RAY? No

Secret JUICE? No

Secret ENERGY? Unlikely

Secret ZIP? No

Secret LOVE? Maybe

Secret CODE? Yes

Secret SAUCE? Yes

**Step 4: Conclusion**

Words that each come after the word “secret”: SANTA AGENT CODE SAUCE

— end of example input and output —

\subsection{Template 24}
Pattern: Puzzle words that each share the same unique characteristic. For this pattern, the connection is with how the puzzle words are structured or how the words sound, not with what the words mean.

Steps to take:

Step 1: Say different puzzle words aloud and try to identify two puzzle words that share a unique characteristic related to how the word is structured or sounds. This is a New York Times puzzle, so unique and particular characteristics are more likely to be the correct connection than generic and ubiquitous characteristics. Think outside the box.

Examples from prior puzzles:

Puzzle words that are each heteronyms: BASS DOVE DESERT WIND

Puzzle words that are each letter spellings: BEE EX GEE JAY

Puzzle words that are each two letters + number: CANINE FREIGHT OFTEN STONE

Puzzle words that are each examples of onomatopoeia: BANG PLOP SPLASH THUD

Puzzle words that are each ”-ough” words that don’t rhyme: BOUGH COUGH DOUGH TOUGH

Puzzle words that are each words you say twice in a row: BOO POM TOM YO

Puzzle words that are each palindromes: BIB EYE GAG POP

Step 2: See if other puzzle words can have that same characteristic. Try to arrive at a group of four puzzle words. Try saying each puzzle word aloud to hear if the linguistic characteristic is present. Take your time and thoroughly consider each possibility. If you can’t identify a group of four words, start over at Step 1.


\end{document}